\documentclass[jair,twoside,11pt,theapa]{article}
\usepackage{jair, theapa, rawfonts, amsfonts, amsmath, amssymb, float, subfig}
\usepackage{mathtools}
\usepackage{bbold}
\usepackage{todonotes}
\usepackage[noend]{algpseudocode}
\usepackage{algorithm,algorithmicx}
\usepackage{mdframed}
\usepackage{url}
\usepackage{booktabs}
\usepackage{paralist}
\usepackage{bm}
\usepackage{listings}
\usepackage{tabularx}
\usepackage{color, colortbl}
\definecolor{Gray}{gray}{0.9}
\usepackage{marvosym}
\usepackage{multirow}
\usepackage{xparse}

\newif\ifarxiv

\arxivtrue

\mdfdefinestyle{plain}{topline=false,bottomline=false,rightline=false,linewidth=2px,linecolor=gray!100,backgroundcolor=gray!20,skipbelow=.5\baselineskip,skipabove=.5\baselineskip}
\mdtheorem[style=plain]{theorem}{Theorem}
\mdtheorem[style=plain]{lemma}[theorem]{Lemma}
\mdtheorem[style=plain]{proposition}[theorem]{Proposition}
\mdtheorem[style=plain]{corollary}[theorem]{Corollary}
\mdfdefinestyle{remark}{topline=false,bottomline=false,rightline=false,linewidth=2px,linecolor=gray!70,backgroundcolor=gray!10,skipbelow=.5\baselineskip,skipabove=.5\baselineskip}

\mdtheorem[style=remark]{example}{Example}
\mdtheorem[style=remark]{definition}{Definition}
\mdtheorem[style=remark]{remark}{Remark}

\DeclareMathOperator*{\argmin}{arg\,min}
\DeclareMathOperator*{\argmax}{arg\,max}
\DeclareMathOperator{\init}{\texttt{init}}
\DeclareMathOperator{\step}{\texttt{step}}
\DeclareMathOperator{\isfinal}{\texttt{is\_final}}
\DeclareMathOperator{\timestep}{\texttt{tstep}}
\DeclareMathOperator{\initcost}{c_{init}}
\DeclareMathOperator{\stepcost}{c_{step}}

\DeclareMathOperator{\sbs}{SBS}
\DeclareMathOperator{\vbs}{VBS}
\makeatletter
\newcommand{\ac}[1]{\expandafter\ifx\expandafter\relax\detokenize{#1}\relax classical AC\else\textnormal{classical AC (#1)}\fi}
\makeatother
\makeatletter
\newcommand{\piac}[1]{\expandafter\ifx\expandafter\relax\detokenize{#1}\relax PIAC\else\textnormal{PIAC (#1)}\fi}
\makeatother
\makeatletter
\newcommand{\dac}[1]{\expandafter\ifx\expandafter\relax\detokenize{#1}\relax DAC\else\textnormal{DAC (#1)}\fi}
\makeatother

\newcommand{\dacsmac}{\dac{SMAC}}
\newcommand{\hydra}{\piac{Hydra}}
\newcommand{\smac}{\ac{SMAC}}
\newcommand{\denselist}{\itemsep -2pt\partopsep 0pt}

\DeclareMathOperator{\formulate}{\texttt{formulate}}
\DeclareMathOperator{\interpret}{\texttt{interpret}}
\makeatother
\let\OldNabla\nabla
\RenewDocumentCommand{\nabla}{e_}{%
    \OldNabla
    \IfValueT{#1}{%
        _{\!#1}
    }%
}
\makeatletter
\newcommand\wordasword[1]{\emph{#1}}
\newcommand{\vect}[1]{\boldsymbol{#1}}

\newcommand{\unaryminus}{\scalebox{0.75}[1.0]{\( - \)}}

\ifarxiv
\else
\jairheading{1}{1991}{1-15}{6/91}{9/91}
\fi

\ShortHeadings{Automated Dynamic Algorithm Configuration}
{Adriaensen, Biedenkapp, Shala, Awad, Eimer, Lindauer, \& Hutter}
\firstpageno{1}

\begin{document}
\title{Automated Dynamic Algorithm Configuration}

\author{\name Steven Adriaensen \email adriaens@cs.uni-freiburg.de \\
\name Andr\'{e} Biedenkapp \email biedenka@cs.uni-freiburg.de\\
\name Gresa Shala \email 
shalag@cs.uni-freiburg.de\\
\name Noor Awad \email  
awad@cs.uni-freiburg.de\\
\addr University of Freiburg, Machine Learning Lab
       \AND
       \name Theresa Eimer \email eimer@tnt.uni-hannover.de\\
       \name Marius Lindauer \email lindauer@tnt.uni-hannover.de\\
       \addr Leibniz University Hannover, Institute for Information Processing
       \AND
       \name Frank Hutter \email  
       fh@cs.uni-freiburg.de\\
       \addr University of Freiburg, Machine Learning Lab \& Bosch Center for Artificial Intelligence
       }


\maketitle

\begin{abstract}
The performance of an algorithm often critically depends on its parameter configuration. 
While a variety of automated algorithm configuration methods have been proposed to relieve users from the tedious and error-prone task of manually tuning parameters, there is still a lot of untapped potential as the learned configuration is \emph{static}, i.e., parameter settings remain fixed throughout the run.
However, it has been shown that some algorithm parameters are best adjusted \emph{dynamically} during execution, e.g., to adapt to the current part of the optimization landscape. Thus far, this is most commonly achieved through hand-crafted heuristics.
A promising recent alternative is to automatically \emph{learn} such dynamic parameter adaptation policies from data. 
In this article, we give the first comprehensive account of this new field of \emph{automated dynamic algorithm configuration (DAC)}, present a series of recent advances, and provide a solid foundation for future research in this field. Specifically, we
(i) situate DAC in the broader historical context of AI research;
(ii) formalize DAC as a computational problem;
(iii) identify the methods used in prior-art to tackle this problem; and
(iv) conduct empirical case studies for using DAC in
evolutionary optimization, AI planning, and machine learning.
\end{abstract}

\section{Introduction}
\label{sec:intro}
Designing robust, state-of-the-art algorithms requires careful design of multiple components.
It is infeasible to know how these components will interact for all possible applications. This is particularly true in the field of artificial intelligence (AI), pursuing ever more general problem-solving methods. This generality necessarily comes at the cost of an increased uncertainty about the problem instances the algorithm will have to solve in practice. To account for this uncertainty, it is common practice to expose difficult design choices as parameters of the algorithm, allowing users to customize them to their specific use case. These algorithm parameters can be numerical (e.g.,  crossover rate or population size in evolutionary algorithms, and the learning rate or batch size in deep learning), but also categorical (e.g., the choice of optimizer in deep learning or the choice of heuristic or search operator in classical planning and meta-heuristics).

\subsection{Algorithm Configuration}
It is widely recognized that appropriate parameter settings are often instrumental for AI algorithms to reach a desired performance level~\shortcite{hutter-cpaior10a,probst-jmlr19a}.
In this paper, we will use the term \wordasword{algorithm configuration} (AC) to refer to the process of determining a policy for setting algorithm parameters as to maximize performance across a problem instance distribution. AC has been widely studied, both in general~\shortcite{birattari-gecco02a,hutter-jair09a,ansotegui-cp09a,hutter-lion11a,lopez-ibanez-orp16a}, as well as in specific research communities~\shortcite{lobo-book07,snoek-nips12a,feurer-bookchapter19}. 

In this work, we focus on a particular kind of AC that is both (i)~automated and (ii)~dynamic. This general framework was recently proposed in a conference paper by \shortciteA{biedenkapp-ecai20a}, and in this article we provide the first comprehensive treatment of the topic. In the remainder of this subsection, we contrast the dynamic/static and automated/manual approaches and position automated dynamic AC as a natural next step.

\paragraph{Dynamic vs. Static AC:}
In \emph{static} AC, parameter settings are \emph{fixed prior to execution}, using the information available at that time, and remain invariant during execution. For example, in evolutionary optimization, the population size is commonly set statically, e.g., as a function of the input dimensionality.
In contrast, in \emph{dynamic} AC (DAC), parameter settings are \emph{varied during execution} using information that becomes available at run time. For example, in machine learning, while static AC would choose a learning rate, possibly dependent on meta-data (e.g., size or modality of the dataset), DAC would propose a learning rate \emph{schedule} that could additionally be a function of time, alignment of past gradients, training/validation losses, etc.
While not all parameters can be varied dynamically, in practice many can, and it often makes sense to do so.
As a general motivating use case, consider parameters that (indirectly) control the exploration/exploitation trade-off: Typically, it makes sense to explore more early on, and to exploit this knowledge in later stages.
Even if the optimal configuration happens to be static, predicting it \emph{upfront} may be very hard, yet the best static configuration may quickly become apparent \emph{while solving} the problem. For instance, if our learning rate is too high, training loss may diverge~\shortcite{bengio-bookchapter12a}.
DAC has been an active research area that has produced various highly practical algorithms leveraging dynamic parameter adaptation mechanisms to empirically outperform their static counter-parts, e.g., Reactive Tabu Search~\shortcite{battiti-orsa94}, CMA-ES~\shortcite{hansen-ec03}, and Adam~\shortcite{kingma-arxiv14a}. Beyond these empirical successes, the potential of DAC has also been shown theoretically~\shortcite{moulines-nips11,senior-icassp13,vanrijn-ppsn18,doerr-toec20,speck-icaps21}.

\paragraph{Automated vs. Manual AC:} The difference between manual and automated AC is \emph{who performs AC}: A human or a machine. Over the last two decades, a variety of general-purpose automated algorithm configurators have been proposed that effectively relieve users from the tedious and time-consuming task of optimizing parameter settings manually~\shortcite{hutter-jair09a,ansotegui-cp09a,kadioglu-ecai10,xu-aaai10a,hutter-lion11a,seipp-aaai15a,lopez-ibanez-orp16a,falkner-icml18a,pushak-gecco20}. However, there is still a lot of untapped potential, as all of these tools perform static AC. In contrast, \emph{dynamic} AC is mostly done manually. Clearly, the human does not directly adjust the parameters during execution; rather, the mechanisms doing this automatically, e.g., learning rate schedules, are products of human engineering. In this work, we will consider deriving such dynamic configuration policies in an automated and data-driven fashion.

\subsection{Summary of Contributions}
In this article, we provide the first comprehensive account of automated DAC. It subsumes and extends four prior conference papers, in which we
\begin{enumerate}
\denselist
    \item established DAC as a new meta-algorithmic framework and proposed solving it using contextual reinforcement learning~\shortcite{biedenkapp-ecai20a};
    \item applied DAC to evolutionary optimization, tackling the problem of step-size adaptation in CMA-ES~\shortcite{hansen-ec03}, and showed that existing manually-designed heuristics can be used to guide learning of DAC policies~\shortcite{shala-ppsn20a};
    \item applied DAC to AI planning, tackling the problem of heuristic selection in FastDownward~\shortcite{helmert-jair06a}, and showed how DAC subsumes static algorithm configuration and can improve upon the best possible algorithm selector~\shortcite{speck-icaps21}; and
    \item presented DACBench, the first benchmark library for DAC, facilitating reproducible results through a unified interface~\shortcite{eimer-ijcai21}.
\end{enumerate}

\noindent{}Here, we go well beyond this previous work, by
\begin{enumerate}[i]
\denselist
\item more thoroughly discussing and classifying related work in different areas (Section~\ref{sec:related_work}), placing recent work on automated DAC in its scientific and historical context;
\item establishing a formal problem formulation (Section~\ref{sec:problem}), offering a novel theoretical perspective on DAC and its relation to existing computational problems;
\item discussing possible methods for solving DAC problems (Section~\ref{sec:methods}), beyond reinforcement learning, and classifying previous work according to their methodology; 
\item extending and using DACBench (Section~\ref{sec:dacbench}) to perform empirical case studies that 
\begin{itemize}
\denselist
\vspace*{-0.1cm}
    \item[-] demonstrate recent successes of automated DAC
    \item[-] provide empirical validation for the benchmark library, and
    \item[-] show that DAC presents a practical alternative to static AC, in various areas of AI: evolutionary optimization (Section~\ref{sec:cmaes}), AI planning (Section~\ref{sec:fastdownward}), and machine learning (Section~\ref{sec:sgd}); and
\vspace*{-0.1cm}
\end{itemize}
\item discussing current limitations of DAC (Section~\ref{sec:conclusion}).
\end{enumerate} 
As such, we provide the first comprehensive overview of automated DAC, a standard reference and a solid foundation for future research in this area.

\section{Related Work}
\label{sec:related_work}
Automated DAC is a new and exciting research area.
However, it did not arise out of thin air, rather it closely relates to, builds on, and tries to consolidate past research efforts. In this section, we place recent work on automated DAC in its scientific and historical context. We start by introducing the terminology we use (Section~\ref{sec:terminology}). Then, we situate DAC in the broader context of AI (Section~\ref{sec:related_areas}). Finally, we discuss some specific prior-art, covering historical work as well as the most recent developments (Section~\ref{sec:prior_art}). 

\subsection{Terminology}
\label{sec:terminology}
Algorithm parameters are omnipresent in computer science. Unsurprisingly, no single set of terms has been consistently used when discussing the problem of how to best set them. In this section, we briefly clarify some of the terms we use, relating them to known alternatives.

We use the term \wordasword{algorithm configuration} (AC) to refer to the process of determining a policy for setting an algorithm's parameters as to maximize performance (or equivalently, minimize cost) across an input distribution.
In the classical AC literature \shortcite{birattari-gecco02a,hutter-jair09a,ansotegui-cp09a}, this process results in a single parameter setting (i.e., a complete assignment of values to parameters) and is called a \wordasword{configuration}.
Later work generalized AC to produce configurations that are a function of the context in which they are used, e.g., the problem instance at hand \shortcite{kadioglu-ecai10,xu-aaai10a}, and most recently the dynamic execution state \shortcite{biedenkapp-ecai20a}. We will use the term \wordasword{configuration policy} to refer to the result of AC in general. To disambiguate the aforementioned AC variants, we add the prefixes \wordasword{ per-distribution} (or also \wordasword{classical}), \wordasword{per-instance} and \wordasword{dynamic}, respectively.
Finally, while AC terminology was introduced in the context of attempts to automate this process, the term itself does not imply automation, i.e., we add prefixes \wordasword{automated} and \wordasword{manual} to specify whether configuration policies are determined automatically or through a manual engineering process, respectively.

In this work, we follow a meta-algorithmic approach to automating AC: We will treat AC as a computational problem to be solved by executing an algorithm. Hence, we have problem instances and algorithms at two different levels and will use the prefixes \wordasword{(D)AC} and \wordasword{target} to disambiguate these:
For example, research on automated DAC aims to find a DAC algorithm for tackling the general DAC problem. In a given DAC problem instance, we aim to find a policy for configuring the parameters of a given target algorithm as to optimize its performance across a distribution of target problem instances. We also use \wordasword{DAC method} and \wordasword{DAC scenario} as a synonym for DAC algorithm and DAC problem instance, respectively.

In machine learning, the problem of setting the hyperparameters of the learning pipeline is known as hyperparameter optimization \shortcite<HPO,> {feurer-bookchapter19}. We consider the more general problem of setting the parameters of \emph{any} target algorithm and therefore adopt a more general terminology~\shortcite{eggensperger-mlj18a}. In meta-learning terms, \wordasword{AC problem solving} corresponds to the \wordasword{outer-loop} and \wordasword{target problem solving} to the \wordasword{inner-loop}.

\begin{figure}[tbh]
    \centering
    \includegraphics[width=\textwidth]{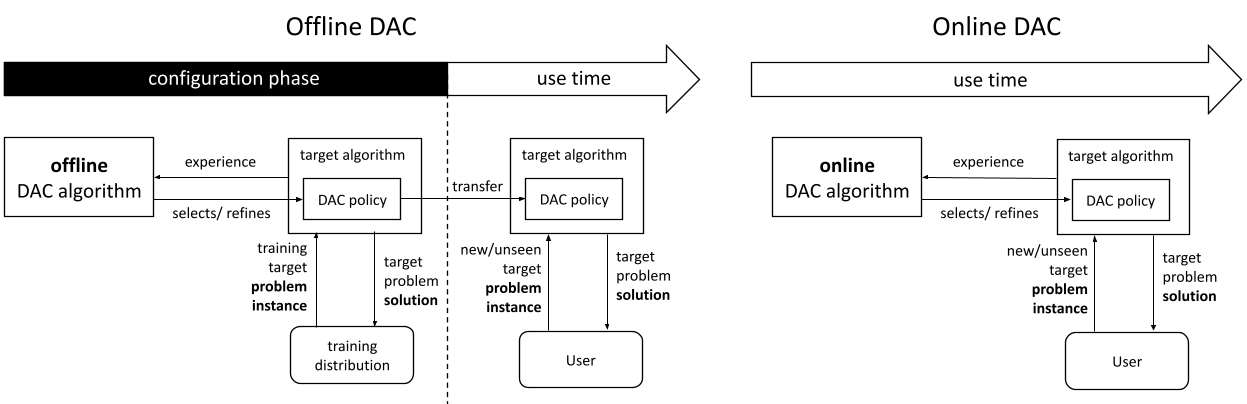}
    \caption{Offline vs. online learning of DAC policies.}
    \label{fig:offline_v_online}
\end{figure}

In heuristic optimization, the terms \wordasword{parameter tuning} and \wordasword{parameter control} are commonly used to refer to static and dynamic algorithm configuration, respectively \shortcite{lobo-book07}.
Also, the terms \wordasword{online} (during use) and \wordasword{offline} (before use) are sometimes used as synonyms for \wordasword{dynamic} and \wordasword{static}, respectively. In this work, we refrain from doing so, reserving these terms to refer to \emph{when (D)AC takes place} (see Figure~\ref{fig:offline_v_online}). In the offline setting, AC takes place in a dedicated \emph{configuration phase} (similar to training in machine learning) where we determine which configuration to use later to solve the problems of actual interest to the user (i.e., at \emph{use time}). In the online setting, AC happens at use time \shortcite{fitzgerald-phd21}. In that sense, offline and dynamic are not mutually exclusive. In fact, most prior-art does DAC offline, determining a dynamic policy offline by using a training set, and at use time simply executing that dynamic policy on new problem instances.

\subsection{Related Research Areas}
\label{sec:related_areas}
While automating DAC is a relatively understudied problem, much research has been performed studying related problems. In what follows, we briefly characterize this work and how it relates to automating DAC. See Appendix~\ref{a:problemtheory} for a more formal treatment of this topic, where we provide problem definitions, possible reductions, and proof their correctness.

\subsubsection{Automated Design of Algorithms / Components}
\label{sec:aad}
The idea of letting computers, rather than humans, design algorithms has been studied in many different communities, using a variety of different methods.
Some well-known, historical examples are \emph{program synthesis}, using logical inference~\shortcite{manna-toplas80}, and \emph{genetic programming}, using evolutionary algorithms \shortcite{koza-book92}. Recent advances in machine learning have prompted a surge in approaches \emph{learning} algorithms, e.g., \emph{Neural Turing machines} \shortcite{graves-arxiv14},  \emph{learning-to-learn}~\shortcite<L2L ,>{andrychowicz-neurips16,lv-icml17,bello-icml17,metz-arxiv20-apalike}, and \emph{learning-to-optimize}~\shortcite<L2O, >{li-iclr17b,kool-iclr18,chen-arxiv21}.

Generally speaking, algorithm parameters can be seen as ``algorithmic design choices'' that are left open at design time. In that sense, automated configuration is naturally viewed as a way of automating \emph{part of} the algorithm design process. This approach has been referred to as ``programming by optimization'' \shortcite<PbO,>{hoos-cacm12}. While previous PbO applications used static AC approaches, the original PbO philosophy envisioned the possibility of varying design decisions at runtime, something naturally achieved by DAC.

A key difference between PbO and the aforementioned design automation approaches is that in PbO algorithms are not designed ``from scratch'', instead only design choices that are ``difficult'' for the human designer are made automatically by the configurator. For instance, DAC aims to design learning rate schedules \shortcite<e.g., >{daniel-aaai16}, but not entire optimizers as in L2L/L2O. 
In summary, DAC can be viewed as automatically designing parameter controlling components, and ``DAC powered PbO'' as a general \emph{semi-automated} algorithm design approach that enables the human designer to bias the design process by embedding prior knowledge (e.g., obtained through decades of algorithmic research), thereby reducing the computational requirements and improving generalization.

\subsubsection{Meta-Algorithmic Frameworks}
\label{sec:meta}

\paragraph{Algorithm Selection}
Problems can be solved using a variety of different algorithms. For example, if we want to sort a sequence of numbers, we could do so using insertion sort, merge sort, quick sort, etc. 
In \emph{algorithm selection}, we determine a mapping from features of the problem instance (e.g., sequence length) to the algorithm best suited to solve it (e.g., that sorts the sequence fastest). While first formalized by \shortciteA{rice76a}, this computational problem only received wide-spread attention two decades later, when it was independently rediscovered by \shortciteA{fink-aips98}, and \shortciteA{leyton-brown-ijcai03} proposed to solve it using machine learning methods. This approach has resulted in various successful applications, e.g., SATzilla \shortcite{xu-jair08a}, a portfolio solver selecting between state-of-the-art SAT solvers to win multiple (gold) medals at the 2007 and 2009 SAT competitions. We refer to \shortciteA{kotthoff-aim14a} and \shortciteA{kerschke-ec19} for surveys on this topic.

\paragraph{Algorithm Scheduling}
It is often difficult to efficiently predict which algorithm will perform best on a given problem instance. In many settings, poor choices may require orders of magnitude longer than optimal choices, and tend to dominate average performance.
In \emph{algorithm scheduling}, instead of selecting a single algorithm, we aim to find an optimal time allocation.
Automated algorithm scheduling was first extensively studied in seminal work by \shortciteA{huberman-science97a} and \shortciteA{gomes-aij01a}, and follow-up work, e.g., by \shortciteA{hoos-tplp14b}, typically focuses on finding a fixed time allocation that works best on average across instances (i.e., per-distribution). These kind of algorithm schedules are also very popular in the AI planning community, e.g., in Fast Downward Stonesoup~\shortcite{helmert-icaps11a}. 
Scheduling has also been combined with algorithm selection to find instance-specific schedules~\shortcite{kadioglu-cp11a,lindauer-lion16a}.
Dynamic scheduling approaches
allocate resources to the algorithms based on runtime information~\shortcite<e.g.,>{carchrae-aaai04,gagliolo-amai06a,nguyen-ial21}. This allows them to exploit the fact that, while it may be difficult to predict which algorithm performs best \emph{in advance}, their relative performance may become apparent early-on in their executions. 
DAC also takes advantage of this property. However, unlike DAC, dynamic scheduling is restricted to allocating resources to \emph{independent} processes; i.e., in scheduling, no information is exchanged between algorithm runs, and resources allocated to all but the one producing the eventual solution are effectively wasted.

\paragraph{Algorithm Configuration}
While algorithm selection chooses between multiple target algorithms on a per-instance basis, classical per-distribution algorithm configuration (AC) is concerned with finding the parameter setting of a single algorithm that performs best across all given instances.
As the space of possible configurations grows exponentially in terms of the number of parameters, research on AC has traditionally focused on (i) efficient search methods, e.g., local search~\shortcite{hutter-jair09a}, genetic algorithms~\shortcite{ansotegui-cp09a} and Bayesian optimization~\shortcite{hutter-lion11a}; and (ii) efficient evaluation of configurations, e.g., using racing~\shortcite{birattari-gecco02a}, adaptive capping~\shortcite{hutter-jair09a}, structured procrastination~\shortcite{kleinberg-ijcai17a} and multi-fidelity optimization~\shortcite{li-jmlr18a}. This line of work has resulted in a variety of automated tools known as \wordasword{configurators} that for any given target algorithm quickly find a configuration that performs well \emph{on average} across a set of target problem instances, e.g., ParamILS~\shortcite{hutter-jair09a}, GGA~\shortcite{ansotegui-cp09a,ansotegui-ijcai15a,ansotegui-sat21a}, SMAC~\shortcite{hutter-lion11a}, iRace~\shortcite{lopez-ibanez-orp16a}, and Golden Parameter Search~\shortcite{pushak-gecco20}; as well as  various theoretical insights~\shortcite{kleinberg-ijcai17a,weisz-icml19a,hall-gecco19,hall-ppsn20}.
Configuration has further been combined with algorithm selection~\shortcite{kadioglu-ecai10,xu-aaai10a}, and algorithm scheduling~\shortcite{seipp-aaai15a}.
However, all of these consider determining a static configuration policy, and the pursuit of similar automated tools and theory for DAC is a natural extension of this line of work.

\subsubsection{Adaptive Operator Selection and Parameter Control}
\label{sec:manual-dac}
\paragraph{Heuristic Approaches}
The potential of varying parameters during execution time is widely recognized and has been extensively studied in various areas of AI. For instance, in heuristic optimization, this problem has been studied in the context of parameter control for evolutionary algorithms \shortcite{aleti-acmcs16a}, reactive search \shortcite{battiti-book08}, and selection hyper-heuristics \shortcite{drake-ejor20}. In machine learning, one hyperparameter that is typically varied is the learning rate, e.g., using global learning rate schedules \shortcite{loshchilov-iclr17a,smith-wacv17a} or adaptive gradient methods \shortcite{kingma-arxiv14a} adopting weight-specific step-sizes.
These works typically consider the dynamic configuration policy as a given and present an empirical/theoretical analysis thereof. Furthermore, the policies themselves were designed by human experts. In contrast, automated DAC is concerned with finding such policies automatically in a data-driven fashion. That being said, prior-art automating DAC does exist and is discussed in Section~\ref{sec:prior_art}. Before doing so, we will briefly discuss a broad class of methods that rely less on human expert knowledge, but that we nonetheless do not generally regard as automated DAC.

\paragraph{Online Learning Approaches}
Many parameter control mechanisms integrate complex feedback loops, learning and optimization mechanisms, creating the potential that the DAC policy is not entirely predetermined by the human, but is rather learned online, while solving the problem instance at hand. All depends on the relative contribution to performance due to (i) the exploration of the hand-crafted DAC algorithm, and (ii) the exploitation of the DAC policy it learns. In an offline setting, distinguishing between (i) and (ii) is easy, as (i) does not occur at test/use time. In online settings, both are intertwined by nature. Note that this does not rule out ``online DAC'', but rather necessitates dedicated analysis that learning indeed takes place. Furthermore, in Section~\ref{sec:dac}, we will define DAC as the problem of finding dynamic configuration policies ``that generalize across a distribution of target problem instances''. Therefore, in our nomenclature, \emph{in order to qualify as automated DAC, an approach must demonstrate the ability to successfully transfer experience across runs of the target algorithm on target problem instances drawn from the same distribution.}
In machine learning terms, automated DAC does not only require learning, but also \emph{meta-learning}. Please note that most previous online learning approaches to parameter control \shortcite<e.g.,>{muller-cec02,carchrae-aaai04,chen-cimca-iawtic06,eiben-esoa06,prestwich-lion07a,wessing-gecco11,di-gaspero-lion12,schaul-icml13,karafotias-gecco14,baydin-iclr18a} trivially do not meet this criterion, as no information is transferred across runs. Note that massive parallel online HPO methods such as Population Based Training \shortcite<PBT, >{jaderberg-arxiv17a-apalike} also fall into this category.

\subsection{Prior-Art: Automated Dynamic Algorithm Configuration}
\label{sec:prior_art}
The term \wordasword{dynamic algorithm configuration} (DAC) was only recently introduced by~\shortciteA{biedenkapp-ecai20a}.
However, various authors had previously (or, in a few cases, concurrently) investigated the possibility of automatically determining policies for varying the configuration of an algorithm on-the-fly. In what follows, we give a brief overview of literature on automated DAC (``avant-la-lettre'').\footnote{We maintain a list of work on automated DAC here:\newline \url{https://www.automl.org/automated-algorithm-design/dac/literature-overview/}} Here, we discuss these by application domain, a methodological overview is presented in Section~\ref{sec:methods}.

Pioneering work by~\shortciteA{lagoudakis-icml00,lagoudakis-endm01a} explored this setting in the context of recursive algorithm selection, observing that sub-problems are better solved using different algorithms (e.g., sorting sub-sequences using different sorting algorithms). While initial results were promising, their approach was limited to recursive target algorithms.
\shortciteA{pettinger-gecco02} considered a more general setting, learning a policy jointly selecting mutation and crossover operators in a genetic algorithm, per generation, based on statistics of the current population.\footnote{
Notably, direct follow-up work by~\shortciteA{chen-cimca-iawtic06}, no longer transferred experience across runs and is therefore not considered prior-art automating DAC (see~Section~\ref{sec:manual-dac}).} 
Various other works have explored automating DAC in the context of genetic algorithms~\shortcite{fialho-amai10,sakurai-sitis10a,andersson-gecco16}, evolutionary strategies~\shortcite{sharma-gecco19}, and heuristic optimization in general~\shortcite{battiti-as12a,lopez-ejor14,ansotegui-aaai17,kadioglu-lion17,sae-dan-gecco20}.
Similar investigations were also conducted in various other communities, e.g., machine learning~\shortcite{daniel-aaai16,hansen-arxiv16,fu-arxiv16,xu-arxiv17,xu-arxiv19-apalike,almeida-arxiv21}, AI planning~\shortcite{gomoluch-icaps19,gomoluch-icaps20}, exact search~\shortcite{bhatia-icaps21}, and quadratic programming~\shortcite{getzelman-acml21,ichnowski-neurips21}.

\shortciteA{biedenkapp-ecai20a} introduced DAC in an attempt to consolidate these isolated efforts and to raise the level of generality in pursuit of algorithms similar to those that exist for static AC. Direct follow-up work has provided additional evidence for the practicality of DAC by learning step-size adaptation in CMA-ES~\shortcite{shala-ppsn20a}, and by learning to select heuristics in the FastDownward planner~\shortcite{speck-icaps21}. These application domains, together with the learning rate control setting from~\shortcite{daniel-aaai16}, have later been released as part of a benchmark suite, called DACbench~\shortcite{eimer-ijcai21}, offering a unified interface that facilitates comparisons between different DAC methods across different DAC scenarios.
In this article, we extend this initial discussion of \shortciteA{biedenkapp-ecai20a} and present a thorough empirical comparison of AC and DAC on these three different real-world DAC applications~\shortcite{daniel-aaai16,shala-ppsn20a,speck-icaps21} using the unified DACbench interface.

\section{Problem Definition}
\label{sec:problem}
In this section, we formalize the computational problem underlying DAC. Here, we first introduce formulations for static AC variants (Section~\ref{sec:static_ac}), and then define the dynamic AC problem (Section~\ref{sec:dac}).

\subsection{Static Algorithm Configuration}
\label{sec:static_ac}
In algorithm configuration, we have some target algorithm $\mathcal{A}$ with parameters $p_1,p_2,\dots,p_k$ that we would like to configure, i.e., assign a value in the domains $\Theta_1,\Theta_2,\dots,\Theta_k$, respectively. Furthermore, we may wish to exclude certain invalid combinations, giving rise to the space of candidate configurations $\Theta \subseteq \Theta_1 \times \Theta_2 \times \dots \times \Theta_k$, called the \emph{configuration space} of~$\mathcal{A}$. In classical per-distribution algorithm configuration, we aim to determine a single $\vect{\theta}^* \in \Theta$ that minimizes a given cost metric $c$ in expectation across instances $i\in I$ of our target problem distribution~$\mathcal{D}$.
This problem can be formalized as follows:
\begin{definition}[Classical / Per-distribution Algorithm Configuration (AC) \label{def:ac}]
Given $\langle \mathcal{A}, \Theta, \mathcal{D}, c \rangle$:
\begin{itemize}
\setlength{\itemindent}{1em}
\item[--] A target algorithm $\mathcal{A}$ with configuration space $\Theta$
\item[--] A distribution $\mathcal{D}$ over target problem instances with domain $I$
\item[--] A cost metric $c: \Theta \times I \rightarrow \mathbb{R}$ assessing the cost of using $\mathcal{A}$ with $\vect{\theta} \in \Theta$ on $i \in I$
\end{itemize}
Find a $\vect{\theta}^* \in \argmin_{\vect{\theta} \in \Theta} \, \mathbb{E}_{i \sim \mathcal{D}} \, [c(\vect{\theta}, i)].$
\end{definition}
In practice, $\mathcal{A}$, $\mathcal{D}$, and $c$ are not given in closed form. Instead, $c$ is typically a black-box procedure that executes $\mathcal{A}$ with configuration $\vect{\theta}$ on a problem instance $i$ and quantifies cost as a function of the desirability of this execution, e.g., how long the execution took, and/or the quality of the solution it found. Note that $\mathcal{D}$ is our true target distribution, i.e., the likelihood $\mathcal{A}$ is presented with an instance $i$ at use time. In the online setting, we are given a sequence of samples from the actual distribution in real time~\shortcite{fitzgerald-phd21}. In the offline setting, we typically do not have access to $i \sim \mathcal{D}$, and are given a set of  instances $I'$ sampled i.i.d. from some representative training distribution $\mathcal{D}' \approx \mathcal{D}$ instead.

Note that unless a single configuration is non-dominated on all instances, 
better performance may be achieved by making the choice of $\vect{\theta}$ dependent on the problem instance $i$ at hand.
This extension is known as:
\begin{definition}[Per-instance Algorithm Configuration (PIAC) \label{def:piac}]
Given $\langle \mathcal{A}, \Theta, \mathcal{D}, \Psi, c \rangle$:
\begin{itemize}
\setlength{\itemindent}{1em}
\item[--] A target algorithm $\mathcal{A}$ with configuration space $\Theta$
\item[--] A distribution $\mathcal{D}$ over target problem instances with domain $I$
\item[--] A space of \emph{per-instance configuration policies} $\psi \in \Psi$ with $\psi: I \rightarrow \Theta$ that choose a configuration $\vect{\theta} \in \Theta$ for each instance $i \in I$.
\item[--] A cost metric $c: \Psi \times I \rightarrow \mathbb{R}$ assessing the cost of using $\mathcal{A}$ with $\psi \in \Psi$ on $i \in I$
\end{itemize}
Find a $\psi^* \in \argmin_{\psi \in \Psi} \, \mathbb{E}_{i \sim \mathcal{D}} \, [c(\psi, i)]$
\end{definition}
Note that the definition above is highly general.
For example, by specifying $\Psi$ accordingly PIAC can put arbitrary constraints on the configuration policies of interest. As a consequence, classical per-distribution AC can be seen as a special case of PIAC only considering constant $\psi$, i.e., $\Psi = \{\psi \,| \, \psi(i) = \psi(i'), \, \forall \, i, i' \in I\}$. More generally, configuration policies could be restricted to be a function of specific features of $i$, or to belong to a specific (e.g., linear) function class. Note that this definition is also strictly more general than \wordasword{unconstrained} PIAC, which is itself a special case. 
Also worth noting is that the cost metric $c$ in this definition can be an arbitrary function of $\psi$ (and $i$). In particular, we do not constrain $c$ to be a function of $\psi(i)$, but allow it to quantify non-functional aspects of $\psi$, e.g., its minimal description length or computational complexity. That being said, most practical PIAC approaches are limited to minimizing $\mathbb{E}_{i \sim \mathcal{D}} \, [c'(\psi(i), i)]$, given some $c': \Theta \times I \rightarrow \mathbb{R}$.

\subsection{Dynamic Algorithm Configuration}
\label{sec:dac}
In dynamic AC, we aim to optimally vary $\vect{\theta} \in \Theta$ while executing $\mathcal{A}$. In order to formalize this problem, we need to define points of interaction where $\mathcal{A}$ can be reconfigured. To this end, we decompose the execution of $\mathcal{A}$ with dynamic configuration policy $\pi \in \Pi$ on problem instance $i \in I$ as shown in Algorithm~\ref{alg:dac}.
Here, we start executing an ``$\init$'' sub-routine bringing $\mathcal{A}$ in some initial state $s \in \mathcal{S}$ only depending on $i$. Subsequently, we iteratively execute ``$\step$'' to determine the next state $s' \in \mathcal{S}$ of $\mathcal{A}$ as a function the current state $s$, instance $i$, and  configuration $\pi(s, i) \in \Theta$. This process continues until $\isfinal(s,i)$ signalling termination and $s$ is returned as solution. When such decomposition $\langle \init, \step, \isfinal \rangle$ is given, we will call $\mathcal{A}$ \emph{step-wise reconfigurable} and define DAC as follows:
\begin{definition}[Dynamic Algorithm Configuration (DAC)]
\label{def:dac}
Given  $\langle \mathcal{A}, \Theta, \mathcal{D}, \Pi, c \rangle$:
\begin{itemize}
\setlength{\itemindent}{1em}
\item[--] A \emph{step-wise reconfigurable} target algorithm $\mathcal{A}$ with configuration space $\Theta$.
\item[--] A distribution $\mathcal{D}$ over target problem instances with domain $I$
\item[--] A space of \emph{dynamic configuration policies} $\pi \in \Pi$ with $\pi: \mathcal{S} \times I \rightarrow \Theta$ that choose a configuration $\vect{\theta} \in \Theta$ for each instance $i \in I$ and state $s \in \mathcal{S}$ of $\mathcal{A}$
\item[--] A cost metric $c: \Pi \times I \rightarrow \mathbb{R}$ assessing the cost of using $\pi \in \Pi$ on $i \in I$.
\end{itemize}
Find a $\pi^* \in \argmin_{\pi \in \Pi} \, \mathbb{E}_{i \sim \mathcal{D}} \, [c(\pi, i)]$
\end{definition}

\begin{algorithm}
  \caption{Step-wise execution of a dynamically configured target algorithm $\mathcal{A}$ \label{alg:dac}}
  \textbf{Input:} Dynamic configuration policy $\pi \in \Pi$; target problem instance $i \in I$ \\
  \textbf{Output:} Solution for $i$ found by $\mathcal{A}$ (following $\pi$)
  \begin{algorithmic}[1]
    \Procedure{$\mathcal{A}$}{$\pi, i$}
      \State $s \gets \init(i)$ \Comment{Initial state by starting the execution of $\mathcal{A}$ on $i$}
      \While{$\lnot \isfinal(s, i)$}
        \State $\vect{\theta} \gets \pi(s, i)$ \Comment{Reconfiguration point: Use $\pi$ to choose next $\vect{\theta}$}
        \State $s \gets \step(s, i, \vect{\theta})$  \Comment{Continue executing $\mathcal{A}$ using $\vect{\theta}$}
      \EndWhile
      \State \Return s \Comment{Execution terminated: Return solution}
    \EndProcedure
  \end{algorithmic}
\end{algorithm}

\noindent Here, we define DAC as a generalization of PIAC, considering configuration policies that do not only depend on $i$, but also the dynamically changing state $s \in \mathcal{S}$ of the target algorithm $\mathcal{A}$, i.e., $\Psi \subseteq \{\pi \, | \pi(i, s) = \pi(i, s'), \, \forall \, s, s' \in \mathcal{S}, \forall \, i \in I\}$. This dynamic state, by definition, provides all information required for continuing the execution of $\mathcal{A}$, however can additionally contain arbitrary features of the execution thus far.
As in PIAC, $c$ can be an arbitrary function of $\pi$ (and $i$). However, often the total cost of executing $\mathcal{A}$ with $\pi$ on $i$ can be decomposed and attributed to the $T$ individual execution steps. Formally: In DAC with \emph{step-wise decomposable cost}, we are given functions $\langle \initcost, \stepcost \rangle$, such that 
\begin{align*}
 c(\pi, i) = \initcost(i) + \sum_{t=0}^{T-1} \stepcost(s_t, i, \pi(s_t, i))
\end{align*}
\begin{align*}
\text{where} \qquad s_t = \begin{cases} 
\init(i) & t = 0 \\  
\step(s_{t-1}, i, \pi(s_{t-1}, i)) & t > 0
\end{cases} \quad \land \quad \isfinal(s_t, i) \Leftrightarrow t = T.
\end{align*}
Note that $\initcost$ and $\stepcost$ only depend on $i$ and $\pi(s_t, i)$, i.e., cannot measure non-functional aspects of $\pi$.

\section{Solution Methods}
\label{sec:methods}
In this section, we discuss methods for solving DAC.
As discussed in Section \ref{sec:manual-dac}, DAC has so far been primarily solved manually, i.e., dynamic configuration policies have been determined by humans and not in an automatic and data-driven way.
In Section~\ref{sec:prior_art}, we discussed previous work exploring \emph{automated} DAC, and in what follows we will give an overview of the methods they used for doing so.
Please note that no dedicated, general DAC solvers exist to date. Instead, prior-art can be viewed as solving DAC \emph{by reduction} to some other well-studied computational problem.\footnote{In Appendix~\ref{a:problemtheory}, we define these related computational problems and discuss reductions more formally.}
Considering the fact that most of this work has been performed in isolation and tackles very different DAC scenarios, the high-level solution approaches followed are remarkably similar.
In particular, we will roughly distinguish between two approaches: ``DAC by reinforcement learning'' (Section \ref{sec:rl4dac}) and ``DAC by optimization'' (Section \ref{sec:opt4dac}), and discuss their relative strengths and weaknesses~(Section~\ref{sec:rl_vs_opt}).

\subsection{DAC by Reinforcement Learning}
\label{sec:rl4dac}
In reinforcement learning \shortcite<RL,>{sutton-book18}, an agent learns to optimize an unknown reward signal by means of interaction with an unknown environment. The RL agent takes actions $a \in A$, observes a transition $T$ from the current state $s \in S$ of the environment to $T(s,a) \in S$, receives a reward $R(s,a) \in \mathbb{R}$, and learns for any state the action maximizing its expected future reward. Formally, the RL agent solves a Markov decision problem $\langle S, A, T, R \rangle$ (MDP, Definition~\ref{def:mdp} in Appendix~\ref{a:rl}).
Here, the transition $T$ and reward $R$ are given in the form of a black box method. Also, the state space $S$ is typically not given explicitly; instead, we are given a procedure for generating initial states and can generate further states using $T$.

\begin{figure}[tb]
    \centering
    \includegraphics[width=\textwidth]{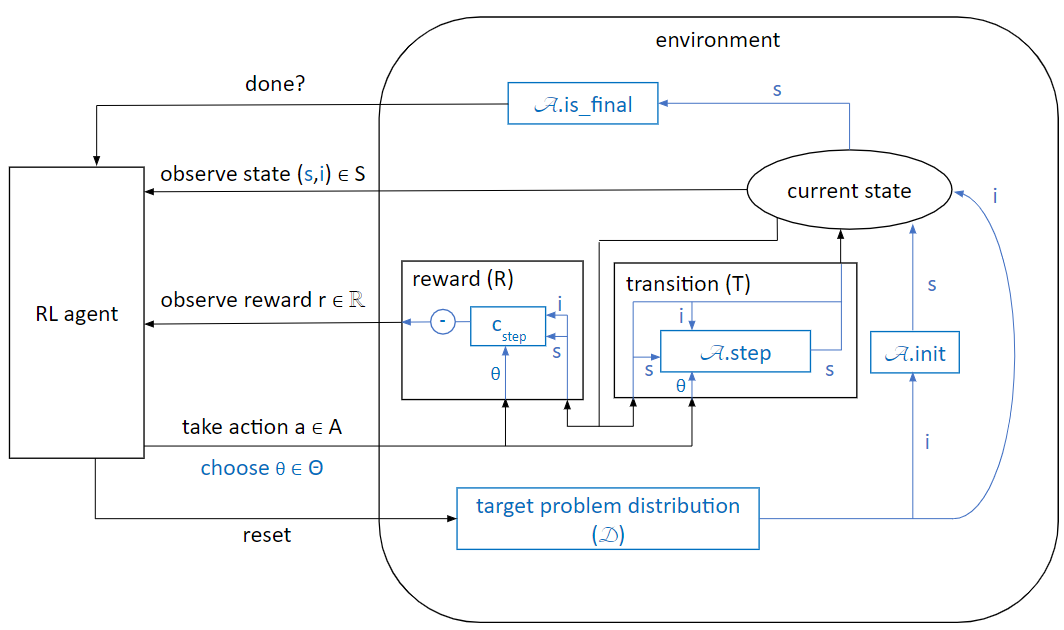}
    \caption{Illustration of DAC by Reinforcement Learning (DAC components in blue)}
    \label{fig:dac_by_rl}
\end{figure}

The RL problem described above is closely related to DAC, and prior art has commonly solved DAC using reinforcement learning methods. In DAC by RL, the environment consists of the target algorithm $\mathcal{A}$ solving some target problem instance $i \in I$. The state of this environment is $s = (s',i) \in S$ with $s' \in \mathcal{S}$ the state of the algorithm, and initial states $(\init(i),i)$ with $i \sim \mathcal{D}$. At every reconfiguration point, the RL agent interacts with this algorithm choosing a configuration $\theta \in \Theta$ as action. The transition dynamics of the environment are fully determined by step-wise algorithm execution, i.e., $T((s', i),\theta) = (\step(s', i, \theta), i)$, and the reward is $R((s', i),\theta) = \unaryminus\stepcost(s', i, \theta)$. See Figure~\ref{fig:dac_by_rl} for an illustration of this approach.
The power of this reduction lies in the fact that the resulting MDP can be solved using the full gamut of existing reinforcement learning methods.

\paragraph{Traditional RL:}
Early \wordasword{DAC by RL} work~\shortcite<e.g.,>{lagoudakis-icml00,lagoudakis-endm01a,pettinger-gecco02,sakurai-sitis10a,battiti-as12a} used
traditional value-based RL methods that learn the optimal state-action value function $Q^*(s,a)$ 
and return the policy $\pi(s) \in \argmax_{a \in A} Q^*(s,a)$. These methods work well when $S \times A$ is small enough to be represented explicitly by a table, but do not scale up. Note that both $S$ and $A = \Theta$ are typically too large in DAC to be modelled in tables.

\paragraph{Modern RL:}
Over the last decade, a series of methodological advances have given rise to a new generation of RL methods that can tackle complex real-world problems \shortcite{mnih-nature13,silver-nature16a,barozet-bioinf20,lee-scirob20},
and that have also been successfully applied to DAC. 
In particular, modern RL methods based on deep neural networks can effectively learn useful representations that allow them to handle complex state and action spaces, using, e.g., 
%
(double) deep Q-learning~\shortcite<DDDDQN, >{hansen-arxiv16,sharma-gecco19,speck-icaps21,bhatia-icaps21}, modern actor critic~\shortcite{andersson-gecco16,xu-arxiv17,ichnowski-neurips21}, and policy gradient methods~\shortcite{daniel-aaai16,xu-arxiv19-apalike,gomoluch-icaps19,shala-ppsn20a,getzelman-acml21,almeida-arxiv21}.

\paragraph{Contextual RL:}
It is worth noting that standard RL methods are not \emph{instance-aware} and will generally not choose their initial state (see Figure~\ref{fig:dac_by_rl}, where $i$ is hidden inside the environment). This is one of the reasons \shortciteA{biedenkapp-ecai20a} proposed to model DAC as a \emph{contextual} MDP~\shortcite<cMDP, >{hallak-corr15}, which consists of a collection of MDPs, one for each instance $i$ (see Definition~\ref{def:cmdp} in Appendix~\ref{a:rl}).
Each MDP $\mathcal{M}(i)$ shares a common action space $S$ and state space $A$ as in traditional RL, but possesses an instance-specific transition function $T_i$ and reward function $R_i$.
This more general formulation allows DAC practitioners to explicitly model variation between instances: Variations in transition dynamics model the differences in target algorithm behaviour between instances (i.e., how the target algorithm progresses in solving an instance) while different reward functions reflect the instance-specific objectives. 
Although a single MDP can capture these dependencies implicitly, the explicit model allows the contextual RL agent to directly exploit this knowledge. For example, instances may vary in difficulty. A contextual RL agent, being aware of different instances and their characteristics, can more easily learn this, allowing the agent to more accurately assign credit for high/low rewards to (i) following a good/poor policy or (ii) solving easy/hard instances. Furthermore, the agent can choose which MDP $\mathcal{M}(i)$ it interacts with, e.g., to gather more experience on harder instances~\shortcite{klink-neurips20,eimer-icml21a}.

\subsection{DAC by Optimization}
\label{sec:opt4dac}
\begin{figure}[tb]
    \centering
    \includegraphics[width=\textwidth]{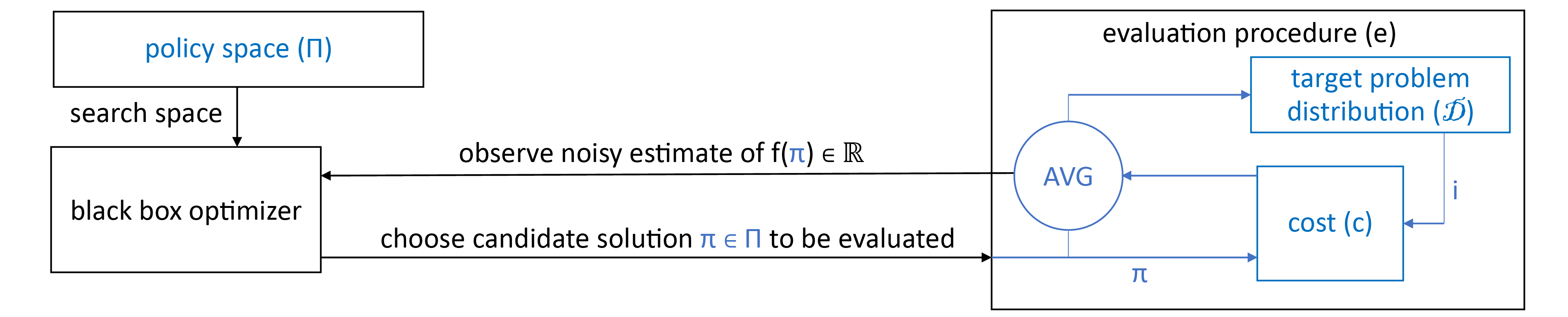}
    \caption{Illustration of DAC by Optimization (DAC components in blue)}
    \label{fig:dac_by_opt}
\end{figure}

Not all prior art automating DAC has done so using reinforcement learning. Instead, some previous works can be viewed as reformulating DAC as a (non-sequential) optimization problem: Given a search space $\Pi$ and an objective function ${f(\pi) = \mathbb{E}_{i \sim \mathcal{D}} \, [c(\pi, i)]}$, find ${\pi^* \in \argmin_{\pi \in \Pi} \, f(\pi)}$. This approach is illustrated in Figure~\ref{fig:dac_by_opt}. 
Optimization covers a wide variety of different methods. In what follows, we give an overview of those used in prior art for ``DAC by optimization'', and distinguish between different variants of optimization depending on (i) search space representation, and (ii) what information about $f$ is used.

\paragraph{Noisy Black Box Optimization:}
In black box optimization (BBO), the only interaction between $f$ and the optimizer is through an evaluation procedure $e$ that returns $f(\pi)$ for any given $\pi \in \Pi$.
A wide variety of black box optimizers exist, specialized for particular kinds of representations.
In the reduction, dynamic configuration policies can be represented in a variety of different ways. For example, prior-art~\shortcite{gomoluch-icaps20} represents policies as real-valued vectors that correspond to the weights of a neural network policy, and optimizes these using evolution strategies. It is worth noting that a similar approach is currently state-of-the-art in learning-to-learn \shortcite{metz-arxiv20-apalike} (see Section \ref{sec:aad}). However, one could go further and also vary the architecture and optimize directly in the space of neural networks, e.g., using methods from neuroevolution~\shortcite{stanley-ec02,stanley-nature19}.
Alternatively, one could follow a symbolic approach, representing policies as programs and use genetic programming \shortcite{koza-book92}. Remark that this freedom comes with responsibility, i.e., making an appropriate choice of representation may be crucial to achieve satisfactory performance.
Next to representation, another difficult choice in this reduction is the evaluation procedure. Since $\mathcal{D}$ is unknown, $e$ cannot evaluate $f$ exactly in general. Instead, we typically evaluate the cost on some finite sample of target problem instances $I' \subseteq I$ with $\forall i \in I': i \sim \mathcal{D}$, and $e(\pi) = \frac{1}{|I'|}\sum_{i \in I'} c(\pi,i)$. However, the choice of $|I'|$ still poses a trade-off between accuracy and cost of evaluation to the DAC by BBO practitioner.

\paragraph{Static Algorithm Configuration:}
We can also solve DAC using classical static algorithm configurators (e.g., SMAC and irace). Assuming we choose a parametric representation $\Lambda$ for the policy space, i.e., $\Pi = \{\pi_{\vect{\lambda}} \, | \, \vect{\lambda} \in \Lambda\}$, the DAC problem can be reformulated as classical AC, where we configure the parameters $\vect{\lambda}$ of the dynamic configuration policy $\pi_{\vect{\lambda}}$, instead of configuring the parameters $\vect{\theta}$ of the target algorithm.\footnote{We proof the correctness of this reduction in Appendix~\ref{a:ac}} While solving DAC using static AC may at first sight seem contradictory, this reduction gives rise to a highly practical solution approach that has been explored extensively in prior art \shortcite{fialho-amai10,lopez-ejor14,andersson-gecco16,ansotegui-aaai17,kadioglu-lion17,sae-dan-gecco20}. An important benefit specific to this approach is that algorithm configurators are \emph{instance-aware} and therefore automate the trade-off between the accuracy/cost of evaluation (using so-called \wordasword{racing} mechanisms), and can even vary $I' \subset I$ dynamically to focus evaluation on those instances providing the most useful information.

\paragraph{Gradient-based Optimization}
In AC, we typically use gradient-free optimization. The motivation is that we cannot generally compute analytical gradients. 
While this is true \emph{in general}, we would like to argue that the specific cases where we can actually compute them are more prominent than one might expect. Assuming a step-wise decomposable cost, we can compute the derivative $\nabla_{\vect{\lambda}} c_i = \frac{\partial c(\pi_{\vect{\lambda}},i)}{\partial \vect{\lambda}}$ from the derivatives of the step-wise cost, the step, and the policy,
using the chain rule (see Appendix~\ref{a:opt}).
When $\stepcost$, $\step$, and $\pi$ can be implemented using the operations in an automated differentiation framework \shortcite<e.g., autograd in Pytorch, >{paszke-neurips17}, these gradients can be calculated efficiently, reliably, without requiring any additional mathematical knowledge from the DAC practitioner.
In fact, in the machine learning community, in particular meta-learning, differentiating through the entire learning process is almost standard practice \shortcite{maclaurin-icml15a,andrychowicz-neurips16,finn-icml17a}.
The potential benefit of this extra piece of information is not to be underestimated. DAC policies may have many hyperparameters, e.g., a neural network with thousands of weights. Gradient-based optimization is an efficient way to navigate extremely high-dimensional spaces, as is evidenced by deep neural networks with millions of parameters being trained almost exclusively using simple first order optimization methods.
That being said, gradients for DAC are no silver bullet.
Computing them, while possible, may require too many computational resources. Furthermore, gradients only provide local information, i.e., an infinitesimal change to every parameter that is guaranteed to reduce cost. When $f$ is particularly rugged, gradients may not provide information about the effect of any reasonably sized change. This phenomena has, in fact, been observed in the context of learning-to-learn~\shortcite{metz-icml19a}.

\subsection{Reinforcement Learning vs. Optimization?}
\label{sec:rl_vs_opt}
Now, we discuss the relative strengths and weaknesses, and argue for the potential of combining both approaches.

\paragraph{Why DAC by RL?} The sequential nature of the problem is arguably the key feature that sets DAC apart from static AC: In static AC, we only have to select a single configuration, while in DAC we must select a sequence of such configurations. RL provides a very general framework for tackling sequential decision problems and was presented as the method of choice for DAC by \shortciteA{biedenkapp-ecai20a}.
DAC by optimization approaches reduce DAC to a non-sequential optimization problem. In doing so, valuable information about the problem is lost that may otherwise be used to solve it more efficiently~\shortcite{adriaensen-ijcai16}.
While executing a target algorithm, an RL agent observes at \emph{every step} what configurations were used, the (immediate) costs this incurred, and how this affected the dynamic state of the algorithm.
In contrast, the same evaluation provides a black box optimizer with a single value (i.e., the sum of costs incurred), at the end of the run.
This inherent relative sample-inefficiency of black box optimization is particularly problematic when target algorithm execution is costly, e.g., takes multiple hours.

\paragraph{Why DAC by Optimization?}
Previous work has shown that optimization can be a practical alternative to RL in simulated environments~\shortcite{mannor-icml03,szita-nc06,salimans-arxiv17a-forapa,chrabaszcz-ijcai18a,majid-techrxiv21}.
While RL aims to exploit sequential information, contemporary RL methods do not always do so successfully. Also, in some scenarios, this information may not add much value, or may even be deceptive (e.g., delayed rewards). Finally, these mechanisms add considerable computational overhead, and complicate implementation. In contrast, optimization methods tend to be simpler, have fewer failure modes, and their often parallel nature makes them well-suited for modern high-performance computing infrastructure. 
Adding to these limitations of RL methods are limitations of the reduction. While DAC is generally reducible to a (noisy) black box optimization problem, the previously discussed reduction to an MDP implicitely assumes (i) the cost function $c$ to be step-wise decomposable an (ii) the space of policies $\Pi$ to be unconstrained. As a consequence, it cannot be used when optimizing non-functional aspects of the policy (e.g., resources it requires to make decisions) or to impose arbitrary hard constraints on $\Pi$ (e.g., which of these $N$ policies is best?).

\paragraph{Beyond RL \emph{or} Optimization:}
Our discussion thus far focused on contrasting both approaches. In what remains, we look at their relation, and argue for the potential of combining them.
First, our ``sequential vs. non-sequential'' discussion can be extended to ``a method's ability to exploit a certain characteristic of DAC'', or not. A good example of a cross-cutting characteristic is \emph{instance-awareness}, both contextual RL and static AC can be viewed as instance-aware extensions of RL and black box optimization, respectively. 
Second, the pitfalls of RL also apply to approaches exploiting other characteristics. For example, gradients in optimization can be similarly deceptive (e.g, exploding/vanishing gradient problem) as immediate rewards. Therefore, while artificially hiding information is useless, blindly relying on it introduces failure modes, and general DAC methods should be carefully designed to only rely on information that is available and useful for the scenario at hand. 
In the context of ``sequential vs. non-sequential'', this suggests the importance of combining reinforcement learning \emph{and} optimization. Further underpinning this conjecture, is the observation that state-of-the-art \emph{static} AC methods combine optimization with machine learning, and reinforcement learning is essentially a dynamic extension of the latter.

\section{Benchmark Library}
\label{sec:dacbench}
In this section, we present DACBench~\shortcite{eimer-ijcai21}, a novel benchmark library for DAC that we will be using in our experiments in Section~\ref{sec:experiments}. We have seen related fields like hyperparameter optimization, static algorithm configuration and algorithm selection profit greatly from focusing on shared benchmark problems~\shortcite{eggensperger-bayesopt13a,hutter-lion14a,bischl-aij16a}.
In these meta-algorithmic domains, standardizing the target algorithm setup did not only increase the accessibility of the field by reducing some of the specialized knowledge required to get started in the field, it also made comparisons between different methods more reliable and reproducible. DACBench provides such a standard for DAC. In what follows, we give a brief overview of the interface it provides, the benchmarks it implements, and prior empirical validation it has undergone. We also discuss novel developments and highlight extensions that were motivated by and/or made specifically in the context of this work.\footnote{A new version of \url{https://github.com/automl/DACBench} (v. 0.1) was released alongside this article.}

\paragraph{Interface:} DACBench builds upon a common RL interface, OpenAI's gym~\shortcite{brockman-arxiv16a-forapa}, as it provides a flexible template for step-wise interaction with the target algorithm. The target algorithm $\init$ is handled in the \texttt{gym.Env.reset} method, with each step-wise interaction handled by the \texttt{gym.Env.step} method.
DACBench extends the \texttt{gym.Env.reset} method to provide the ability to select the problem instance $i$ to be solved. Listing~\ref{lst:dacgym} shows how DAC components are mapped onto the gym interface in the benchmarks. These essentially implement the DAC by contextual RL reduction, discussed in Section~\ref{sec:rl4dac}. The result is a simple-to-use interface, allowing DAC researchers to work across application domains, without requiring domain expertise, and providing an easy-to-use template for applying DAC to new domains. While the interface is modelled after the RL formulation of DAC, it can be used with a variety of approaches described in Section~\ref{sec:opt4dac}.
That being said, the original DACbench interface strongly focused on conventional RL. In the scope of this work, we have extended the interface from the first release of DACBench. In accordance with our proposed definition of DAC, we have taken a broader perspective beyond standard RL, and made various interface changes to provide better support for alternative approaches. For example, users can now specify rich structured configuration spaces as opposed to the simplistic action spaces supported by conventional RL methods. Directly controlling instance progression is easier now as well, providing 
a better base for developing instance-aware solution methods for DAC.

\lstset{escapeinside={<@}{@>}}
\begin{lstlisting}[language=Python,caption={A generic python implementation of a gym environment using components of a DAC scenario (in blue) with decomposable cost and an input-constrained policy space $\Pi_\phi$. Practical DACBench benchmarks implement a similar mapping, but DAC components are typically not strictly separated, e.g., $\mathcal{A}.\step$ and $\stepcost$ would typically be calculated jointly. Note that the \texttt{gym.Env.step} method, despite its name, does far more than merely computing $\mathcal{A}.\step$: It implements the transition dynamics ($T$) and reward signal ($R$). Furthermore, unlike $\mathcal{A}.\step$, it is stateful, does not take the state $(s,i)$ as input, and does not necessarily return the new state. Instead, it more generally returns what is called an observation $\phi(s,i)$ which may \emph{abstract} arbitrary aspects of the internal state, i.e.,  DACBench technically reduces DAC to a contextual \emph{partially observable} MDP (cPOMDP). Note that the learned policy in POMDPs is a function of the \emph{observable} state, and hence $\phi$ can be viewed as modeling a policy space constraint of the form ${\Pi_\phi = \{\pi \, | \, \phi(s,i) = \phi(s',i') \implies\ \pi(s,i) = \pi(s',i')\}}$.},captionpos=b,label={lst:dacgym}]
class DACEnv(gym.Env):

    def __init__(self, <@\textcolor{blue}{$\mathcal{A}$}@>, <@\textcolor{blue}{$\Theta$}@>, <@\textcolor{blue}{$\mathcal{D}$}@>, <@\textcolor{blue}{$\Pi_\phi$}@>, <@\textcolor{blue}{$\stepcost$}@>):
        self.<@\textcolor{blue}{$\mathcal{A}$}@>, self.<@\textcolor{blue}{$\mathcal{D}$}@>, self.<@\textcolor{blue}{$\stepcost$}@>, self.<@\textcolor{blue}{$\phi$}@> = <@\textcolor{blue}{$\mathcal{A}$}@>, <@\textcolor{blue}{$\mathcal{D}$}@>, <@\textcolor{blue}{$\stepcost$}@>, <@\textcolor{blue}{$\Pi_\phi.\phi$}@>
        self.action_space = <@\textcolor{blue}{$\Theta$}@>

    def reset(self, <@\textcolor{blue}{i}@>=sample(self.<@\textcolor{blue}{$\mathcal{D}$}@>)):
        <@\textcolor{blue}{s}@> = self.<@\textcolor{blue}{$\mathcal{A}$.$\init$}@>(<@\textcolor{blue}{i}@>)
        self.state = (<@\textcolor{blue}{s}@>, <@\textcolor{blue}{i}@>)
        return self.<@\textcolor{blue}{$\phi$}@>(<@\textcolor{blue}{s}@>, <@\textcolor{blue}{i}@>)
        
    def step(self, <@\textcolor{blue}{$\theta$}@>):
        <@\textcolor{blue}{s}@>, <@\textcolor{blue}{i}@> = self.state
        <@\textcolor{blue}{s}@> = self.<@\textcolor{blue}{$\mathcal{A}$.$\step$}@>(<@\textcolor{blue}{s}@>, <@\textcolor{blue}{i}@>, <@\textcolor{blue}{$\theta$}@>)
        r = self.<@\textcolor{blue}{$\stepcost$}@>(<@\textcolor{blue}{s}@>, <@\textcolor{blue}{i}@>, <@\textcolor{blue}{$\theta$}@>)
        done = self.<@\textcolor{blue}{$\mathcal{A}$.$\isfinal$}@>(<@\textcolor{blue}{s}@>, <@\textcolor{blue}{i}@>)
        self.state = (<@\textcolor{blue}{s}@>, <@\textcolor{blue}{i}@>)
        return self.<@\textcolor{blue}{$\phi$}@>(<@\textcolor{blue}{s}@>, <@\textcolor{blue}{i}@>), r, done, None
 
\end{lstlisting}

\paragraph{Benchmarks:} An overview of the benchmarks currently included in DACBench is given in Table \ref{tab:dacbenchmarks}.
It includes several benchmarks that we have either added in the latest release or at least improved significantly. The original SGD-DL benchmark (see Section~\ref{sec:sgd} for a thorough description) was extended to mimic the experimental setup from~\shortciteA{daniel-aaai16} as closely as possible. The CMA-ES benchmarks (CMAStepSize and ModCMA) are now based on IOHProfiler~\shortcite{doerr-arxiv18} and thus provide a DAC interface for a well-known and important tool in the EA community.\footnote{The original \emph{pycma} version of CMAStepSize is still supported and used in Section~\ref{sec:cmaes}.}
TheoryBench is a completely new benchmark, published by~\shortciteA{biedenkapp-gecco22}, where one is to dynamically configure the mutation rate of a (1+1) random local search algorithm for the LeadingOnes problem. This is a particularly interesting setting as the exact runtime distribution is very well understood in this setting \shortcite{doerr-tcs19}. In particular, it is possible to compute optimal dynamic configuration policies for various different problem sizes and configuration spaces.
Finally, a continuous Sigmoid variation and SGD on polynomials provide additional artificial benchmarks for efficient evaluation of DAC algorithms.

\paragraph{Empirical Validation:}
DACBench is a very recent library. As a consequence, it has not yet been used in prior-art. \shortciteA{eimer-ijcai21} focused on providing a unified interface to a variety of benchmarks and analyzed specific properties of these benchmarks based on the behavior of static policies and simple hand-crafted dynamic policies. Here, we describe the first applications of practical DAC methods to these benchmarks, and provide important empirical validation for DACBench.

\begin{table}[tbhp]
    \centering
    \resizebox{\textwidth}{!}{%
    \begin{tabular}{cccp{14cm}}
        \toprule
        Benchmark & Domain & Status & Description \\
        \midrule
         Sigmoid & Toy & Extended & Control $k$ parameters to trace a different sigmoids each~\shortcite{biedenkapp-ecai20a}. \\ 
         Luby & Toy & Original & Select the correct next term in a shifted luby sequence~\shortcite{biedenkapp-ecai20a}. \\
         CMAStepSize & EA & Extended & Control the step size in CMA-ES~\shortcite{shala-ppsn20a}. \\
         FastDownward & Planning & Original & Control heuristic selection in FastDownward~\shortcite{speck-icaps21}. \\
         SGD-DL & DL & Extended & Control the SGD for neural network training~\shortcite{daniel-aaai16}.\\
         TheoryBench & EA & New & Control the mutation rate of (1+1)RLS for LeadingOnes~\shortcite{biedenkapp-gecco22}.\\
         ModCMA & EA & New & Control design choices (e.g., base sampler used) of CMA-ES~\shortcite{vermetten-gecco19}. \\
         ToyGD & Toy & New & Control the learning rate of gradient descent on polynomial functions. \\
         \bottomrule
    \end{tabular}%
    }
    \caption{DACBench Benchmarks. ``Status'' compares the current state of each benchmark to the benchmarks originally introduced by  \shortciteA{eimer-ijcai21}.}
    \label{tab:dacbenchmarks}
\end{table}

\section{Empirical Case Studies}
\label{sec:experiments}
In this section, we discuss in more detail three successful applications of automated DAC in different areas of AI: evolutionary optimization~\shortcite[Section~\ref{sec:cmaes}]{shala-ppsn20a}, AI planning~\shortcite[Section~\ref{sec:fastdownward}]{speck-icaps21}, and machine learning~\shortcite[Section~\ref{sec:sgd}]{daniel-aaai16}.
The primary purpose of this section is to complement the general, big picture discussions in previous sections with some concrete practical examples of automated DAC. Here, we cover our own work in this area~\shortcite{shala-ppsn20a,speck-icaps21}, supplemented with a machine learning application~\shortcite{daniel-aaai16} for diversity.
In these case studies, we also conducted additional experiments to answer the following research questions.

\paragraph{RQ1: Can we reproduce the main results of the original papers using DACBench?} \hfill
Since it is well known that RL results are hard to reproduce~\shortcite{henderson-aaai18a}, in order to provide a solid foundation for experimental work in the field we believe it to be important to repeat the original experiments, this time using the publicly available re-implementations provided by DACBench (i.e., the CMAStepSize, FastDownward, and SGD-DL benchmarks) and to compare the results obtained to those of the original papers.
Beyond insights into the reproducibility of the prior work, this analysis provides empirical validation for DACBench: This is the first study investigating whether, and to what extent, the benchmarks in DACBench permit reproducing the original results.
Further, it is worth noting that the work by~\shortciteA{daniel-aaai16} is closed source, and that this is the first reproduction of their experiments with open-source code.

\paragraph{RQ2: Does DAC outperform static AC in practice?} \hfill \\
Theoretically, an optimal DAC policy will be at least as good as an optimal static AC policy.
In practice, however, the superiority of DAC is not guaranteed, since practical DAC methods may not be capable of finding an optimal/better DAC policy and/or doing so may require more computational resources than available.
To investigate this, for each scenario in our case studies, we compare the anytime performance of the DAC method used to that of static AC baselines: We run SMAC~\shortcite<as a classical AC method, >{hutter-lion11a,lindauer-jmlr22a} and Hydra\footnote{Hydra combines SMAC with an algorithm selection method of choice. Since most of the considered benchmarks do not have instance features, we will assume an oracle selecting the best configuration in the portfolio. We treat the maximum size of the portfolio as a case study dependent hyperparameter and detail this choice in the respective experimental setups.}~\shortcite<as a PIAC method, >{xu-aaai10a} on the same problem, and compare the performance of the best dynamic/static policies found at any time during the configuration process. We further include the theoretical upper bounds for classical AC ($\sbs = \min_{\vect{\theta}\in\Theta}\frac{1}{|I'|}\sum_{i \in I'} c(\vect{\theta}, i)$) and PIAC ($\vbs = \frac{1}{|I'|}\sum_{i \in I'}\min_{\vect{\theta}\in\Theta} c(\vect{\theta}, i)$) as reference, to distinguish practical from inherent limitations of static AC.\footnote{Note that the acronyms SBS (single best solver) and VBS (virtual best solver) stem from the algorithm selection literature. More details on how these theoretical bounds were determined can be found in the experimental setup of the respective case studies.}

In what follows, we discuss our three case studies, in each case presenting an introduction to the domain, the problem formulation as an instance of DAC, the solution method, the experimental setup, the results, and a discussion thereof.\footnote{Code for reproducing these experiments is publicly available: \newline \url{https://github.com/automl/2022_JAIR_DAC_experiments}}

\subsection{Step Size Adaptation in CMA-ES}
\label{sec:cmaes}
The first problem we consider is to dynamically set the step-size parameter of the Covariance Matrix Adaptation Evolution Strategy \shortcite<CMA-ES,>{hansen-ec03}, an evolutionary algorithm for continuous black box optimization.
Each generation $g$, CMA-ES evaluates the objective value $f$ of $\lambda$ individuals $x_{1}^{(g+1)},...,x_{\lambda}^{(g+1)}$ sampled from a non-stationary multivariate Gaussian distribution $\mathcal{N}(\vect{\mu}^{(g)}, {\sigma^{(g)}}^2 \cdot C^{(g)})$. Then, based on the outcome of these evaluations, CMA-ES heuristically adapts the parameters $\vect{\mu}$, $\sigma$, $C$ of the search distribution aiming to increase the likelihood of generating better individuals next generation. In particular, the step-size parameter $\sigma$ controls the scale of the search distribution and CMA-ES by default adjusts it using Cumulative Step Length Adaptation \shortcite<CSA,>{hansen-cec96}.
CSA is a hand-designed heuristic and thus implicitly makes assumptions about the properties of the tasks it is applied on. In \shortciteA{shala-ppsn20a}, we investigated the possibility of learning step-size adaptation in a data-driven fashion, optimized for the task distribution at hand, i.e. automated DAC. 

\paragraph{Problem Formulation:} Below, we briefly detail each of the DAC components:

\begin{description}
\item[$\mathcal{A}$, $\Theta$:] The target algorithm to configure in this scenario is CMA-ES. As in \shortciteA{shala-ppsn20a}, we use the \emph{pycma} distribution of CMA-ES. Its interface allows for step-wise execution of CMA-ES. CMA-ES is initialized with a given initial mean $\vect{\mu}^{(0)}$ and step-size $\sigma^{(0)}$ ($C^{(0)} = \mathbb{1}$). Each generation $g$, we 
\begin{enumerate}
\item sample $\lambda$ individuals $x_{1}^{(g+1)},...,x_{\lambda}^{(g+1)}$ from $\mathcal{N}(\vect{\mu}^{(g)}, {\sigma^{(g)}}^2 \cdot C^{(g)})$
\item evaluate the objective function values $f(x_{1}^{(g+1)}),...,f(x_{\lambda}^{(g+1)})$ of these individuals
\item adapt the distribution parameters $\vect{\mu}^{(g+1)}$, $\sigma^{(g+1)}$, $C^{(g+1)}$ for the next generation.
\end{enumerate}
In this final step, the mean $\vect{\mu}$ and covariance $C$ are adapted as usual in CMA-ES, while the step-size $\sigma$ is to be reconfigured dynamically in the range $\Theta = \mathbb{R}^+$.
\item[$\mathcal{D}, I$:] Instances correspond to tuples consisting of a black box function $f$ and an initial search distribution. Here, the latter is isotropic and defined by an initial mean $m^{(0)}$ and step-size $\sigma^{(0)}$.

\item[$\Pi$]: The policies are constrained to be functions of a specific observable state composed of: 
\begin{inparaenum}[(i)]
\item the current step-size value $\sigma^{(g)}$;
\item the current cumulative path length $p_{\sigma}^{(g)}$~\shortcite{hansen-cec96};
\item the history of changes in objective value, i.e., the normalized differences between successive objective values, from \emph{h} previous iterations; and
\item the history of step-sizes from \emph{h} previous iterations.
\end{inparaenum}

\item[$c$]: The cost metric used is ``the likelihood of outperforming CSA''. Assuming we perform two runs of CMA-ES, one using $\pi$, the other CSA, it measures how likely the latter is to obtain a better final solution than the former. We estimate this probability based on pairwise comparisons of $n = 25$ runs varying only the random seed of CMA-ES, i.e.,
\begin{equation*}
\label{eq:metric}
    c(\pi, i) = \frac{\sum_j^n\sum_k^n \mathbb{1}_{\pi_j<CSA_k}}{n^2}
\end{equation*}
where $\mathbb{1}_{\pi_j<CSA_k}$ is the function indicating that our policy resulted in a lower final objective value than the baseline using CSA, when comparing runs $j$ and $k$. Note that a benefit of this cost metric is that it is easy to interpret, both conceptually and in terms of statistical significance. As explained in more detail in the original publication \shortcite[Appendix C]{shala-ppsn20a}, it has a direct correspondence with the Wilcoxon rank sum statistic. For $n = 25$, estimates $c(\pi, i) \geq 0.64$ imply $\pi$ significantly outperformed CSA (at 95\% confidence, one-sided).
\end{description}

\paragraph{Solution Method:}
In \shortciteA{shala-ppsn20a}, we proposed to use existing hand-crafted heuristics to warm-start DAC. To this end, we adopted the methodology proposed by \shortciteA{li-iclr17b} in the context of L2O and used guided policy search \shortcite<GPS,>{levine-neurips14}, a reinforcement learning method originating from the robotics community. In GPS, a teacher (typically a human) provides some initial sample trajectories that the RL agent first learns to imitate and then iteratively refines without further interaction with the teacher. To learn step-size adaptation policies, in \shortciteA{shala-ppsn20a}, we used CSA as a teacher and extended GPS with \wordasword{persistent teaching}, meaning that at each iteration the GPS agent obtains a fixed fraction (the \emph{sampling rate}, a hyperparameter) of its sample trajectories from the teacher, encouraging it to continually learn from CSA. Folowing \citeauthor{li-iclr17b}, the area under the curve (AUC) was used as a reward signal for GPS, instead of negated cost. Here, the reward at step $t$ is $-\min_{x \in X_t} f(x)$ where $X_t = \{x_i^{(g)} \, | \, g \leq t \}$ is the set of individuals evaluated up until step $t$. This reward signal, unlike negated cost, is dense and actively encourages learning policies with good anytime performance.

\paragraph{Experimental Setup:}
In our experiments, we used the DACBench implementation of 
the CMAStepSize benchmark. Replicating the original setup, we set population size $\lambda=10$, history length $h=40$, terminate CMA-ES after 50 generations, and model policies as fully connected feed-forward neural networks having two hidden layers with 50 hidden units each and ReLU activations. Note that in \shortciteA{shala-ppsn20a}, we considered a collection of different scenarios varying in target distribution: 
\begin{inparaenum}[(i)]
    \item single black box function, different initial search distributions;
    \item black box functions of the same type, but different dimensionalities and initial search distributions; and
    \item black box functions of different types and initial search distributions.
\end{inparaenum}
In our case study here, we only reproduce and discuss the results for the third scenario, as it considers learning policies that generalize across different black box functions. Here, the training setup consists of $100$ training instances: $10$ different black box functions, with $10$ different initial search distributions each. For testing, $12$ other black box functions were used with a specific initial search distribution. In both cases, the  functions used were taken from the BBOB-2009 competition~\shortcite{hansen-tech09}.
We perform five independent GPS training runs using the original hyperparameters, each performing a total of 40000 CMA-ES runs and taking 8-10 CPU hours on our system. In our comparison of anytime performance to static AC, the same budget was used for \smac{} and \hydra{}. A maximum portfolio size of 10 was used for Hydra.
To determine $\sbs$ and $\vbs$, we discretized $\Theta$ (1000 values equidistant in [0.1, 2.0]) and evaluated $c(\vect{\theta},i)$ for all (1000 $\times$ 100) combinations of $\vect{\theta} \in \Theta$ and $i\in I'$.

\paragraph{Results:}
\begin{figure}%
    \centering
    \includegraphics[width=0.75\textwidth]{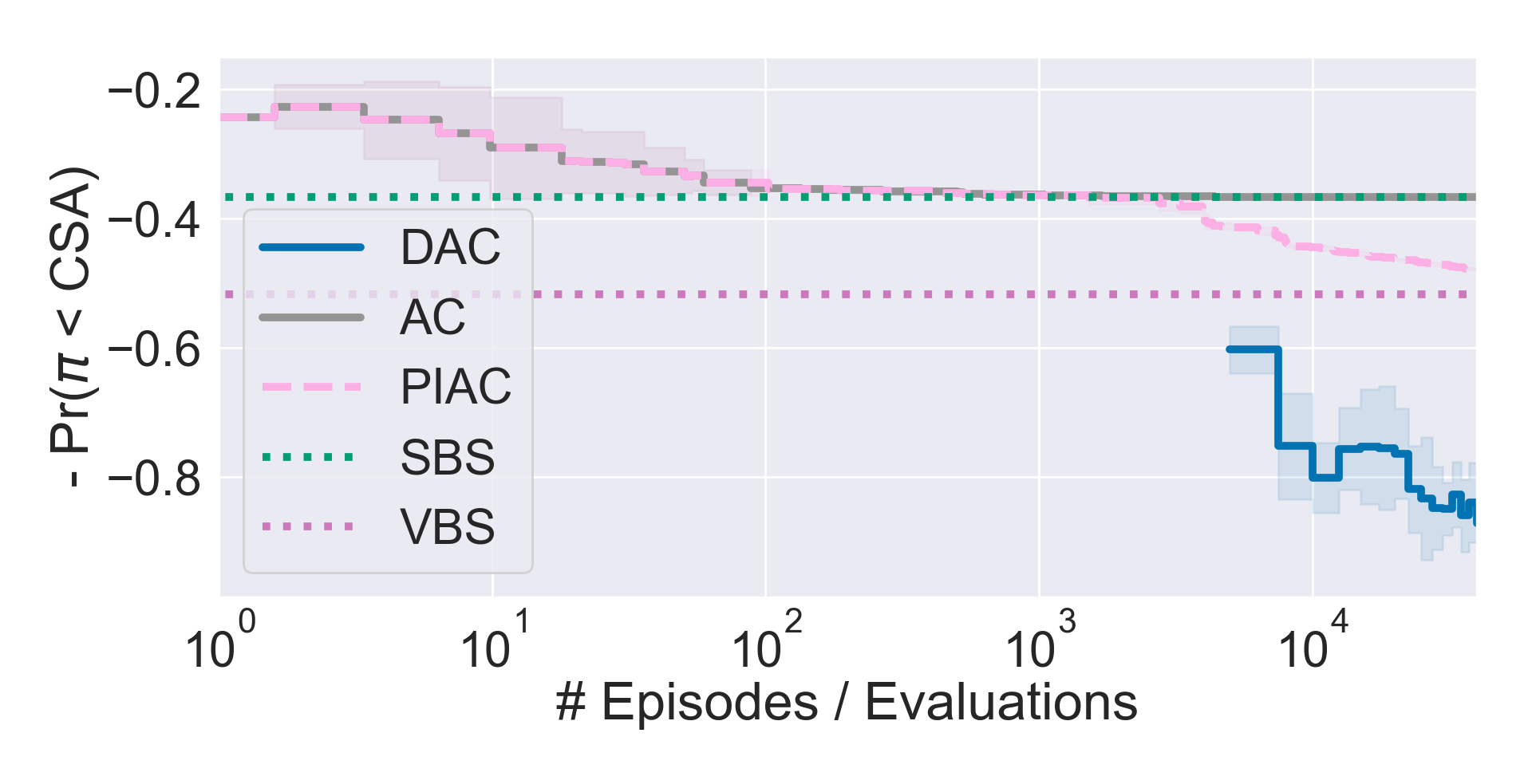}%
    \caption{Incumbent performance of \dac{GPS}, \hydra{}, and \smac{} when determining a step-size configuration policy for CMA-ES. Solid lines depict the mean of five independent configuration runs and the shaded area the standard deviation. SBS depicts the single best configuration and VBS the oracle configuration portfolio across all instances.}%
    \label{fig:cma_dac_vs_ac}%
\end{figure}
Figure~\ref{fig:cma_dac_vs_ac} compares the anytime training performance of \dac{GPS} to that of \smac{} and \hydra{} when learning step-size adaptation.
Classical AC and \piac{} initially show similar anytime behavior, where the former reaches $\sbs$ performance after 1000 evaluations, the latter further improves, but does not reach $\vbs$ performance within the given budget of 40000 evaluations.
In contrast, \dac{GPS} has a minimum budget of 5000 evaluations, however, the initial incumbent immediately outperforms the $\vbs$ and further improves to eventually find a DAC policy that on average outperforms CSA on 87\% of the runs on the training setting. 
\begin{figure}
    \centering
    \includegraphics[width=\textwidth]{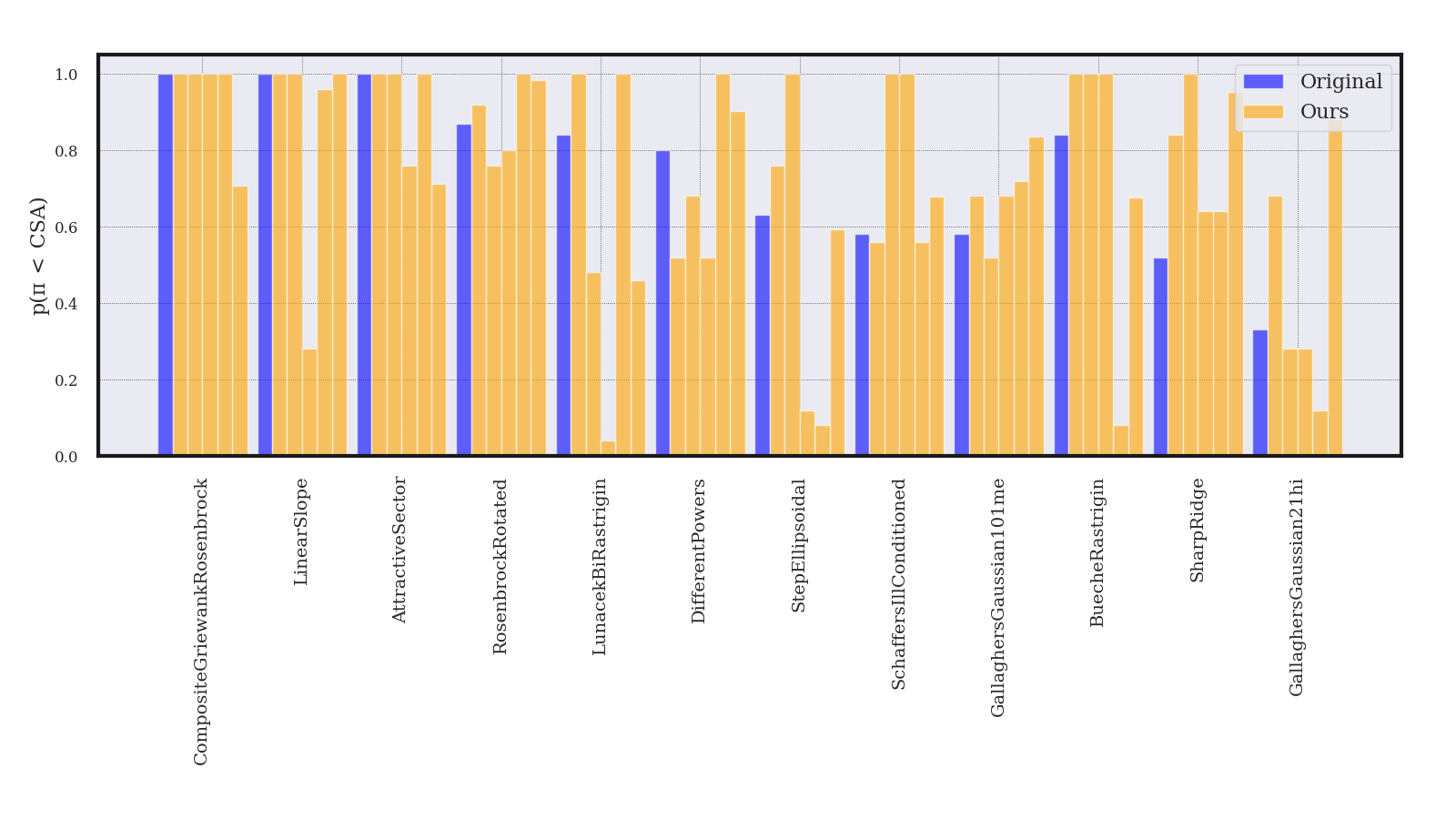}
    \caption{Likelihood of the policies learned by GPS (for five runs) outperforming CSA on $12$ unseen test functions. The reported values from \shortcite{shala-ppsn20a} are shown in blue, whereas the results for the five learned policies are shown in yellow, in a consistent order.}
    \label{fig:bar_13bbob}
\end{figure}
Figure~\ref{fig:bar_13bbob} shows the likelihood of the five learned policies outperforming CSA on the $12$ unseen test functions. Here, for each of five individual seeds, we observe that the learned policies significantly (p($\pi<$ CSA) $\geq 0.64$, $\alpha = 0.05$) outperformed CSA on $10, 9, 10, 8, 8$ of the 12 unseen test functions, while being significantly outperformed on $0, 1, 0, 3, 3$, respectively.
In comparison to the original, the learned policies performed similarly when averaging costs across all test functions/policies (0.74 vs 0.75 originally). However, it is worth noting that the average performance of the individual policies and the performance profile across the test functions varies more strongly.

\paragraph{Discussion:}
On a high level, we could reproduce our results from~\shortciteA{shala-ppsn20a}, showing that the learned policies for step-size adaptation can outperform CSA even on functions not seen during training. Since the DACBench implementation, to the best of our knowledge, exactly replicates the original setup, we assume the observed differences to be a consequence of variability across training runs. This is supported by our observation that the five different runs of GPS (varying only in random seed) resulted in policies whose test cost ranged from -0.83 to -0.65 (vs. -0.75 originally).
Our analysis of the anytime performance revealed another weakness of the approach: Its relatively high up front cost. It is worth noting that this cost includes the teacher runs ($25 \times 100$ runs using CSA) we performed to warm-start GPS. Nonetheless, since GPS maintains an independent controller per instance, its computational cost will generally scale linearly with the number of training instances. Further, it is difficult to predict in advance how many training instances and runs per training instance suffice. In comparison, the static approaches in our comparison follow a more incremental approach resulting in a better anytime performance. That being said, the best static policy did not significantly outperform CSA. As such, independent of the specific approaches, our results provide further evidence of the importance of dynamic step-size adaptation, showing that DAC policies (learned, but also CSA) are competitive with and/or outperform their static counterparts, even on relatively short CMA-ES runs.

\subsection{Heuristic Selection in FastDownward}
\label{sec:fastdownward}
Heuristic search is one of the most widely used and successful approaches to AI planning.
This type of search makes use of heuristics to estimate the distance to some desired goal state as a cheap proxy of having to directly evaluate the true distance.
Over decades of research, many different heuristics have been developed for a variety of problem domains.
No single heuristic works best on all problem instances~\shortcite{wolpert-macready-tr1995}.
Thus, the AI planning community has made use of meta-algorithmic approaches such as algorithm selection, algorithm scheduling and algorithm configuration~\shortcite{helmert-icaps11a,seipp-ipc14a,fawcett-icaps14a,seipp-aaai15a,sievers-aaai19}.
However, one limiting factor of these approaches is that they do not take the internal dynamics of the planning system into account and only adapt to a set of problem instances (per-distribution) or individual problem instances (per-instance).
It has been shown that using hand-crafted policies to switch between heuristics to adapt to changing conditions can greatly improve performance~\shortcite{richter-icaps09,roeger-icaps10}.
\shortciteA{speck-icaps21} proposed to use dynamic algorithm configuration (DAC) to automatically determine a policy that selects at each individual planning step which heuristic to follow, out of a set of heuristics sharing their progress.
That work showed that DAC is in theory able to outperform prior meta-algorithmic approaches and empirically validated this by outperforming the theoretical best algorithm selector (a.k.a. virtual best solver) on multiple domains. 
Relatedly, \shortciteA{gomoluch-icaps19,gomoluch-icaps20} previously investigated automated DAC in the context of switching between different search strategies in AI planning. To provide an exemplary showcase of the potential of DAC in AI planning, we focus on the heuristic selection problem here.

\paragraph{Problem Formulation:} Below, we briefly detail each of the DAC components:
\begin{description}
\item[$\mathcal{A}$, $\Theta$:]
The target algorithm to configure in this scenario is the popular FastDownward Planner \shortcite{helmert-jair06a}.
To make it step-wise executable, and to allow communication with a dynamic configuration policy, \shortciteA{speck-icaps21} proposed to set up a socket communication such that DAC can change heuristics after each node expansion.
The configuration space consists of four heuristics\footnote{In an additional experiment, \shortciteA{speck-icaps21} showed that even with an increased action space, including the landmark-count heuristic \shortcite{richter-et-al-aaai2008}, DAC was still capable of learning better policies than the considered baselines.
Here, we limit ourselves to the original configuration space which only includes four heuristics.} (i.e., a single categorical  parameter), commonly used in satisficing planning:
\begin{inparaenum}[(i)]
    \item the FF heuristic $h_{\text{ff}}$~\shortcite{hoffmann-nebel-jair2001},
    \item the causal graph heuristic $h_{\text{cg}}$~\shortcite{helmert-icaps2004},
    \item the context-enhanced additive heuristic $h_{\text{cea}}$~\shortcite{helmert-geffner-icaps2008}, and
    \item the additive heuristic $h_{\text{add}}$~\shortcite{bonet-geffner-aij2001}.
\end{inparaenum}
The planning system is terminated when a solution is found. Since some runs may fail to find a solution, \shortciteA{speck-icaps21} also limited the maximal run length. During the configuration phase (training of the RL agent) an individual solution attempt can run for at most $7500$ steps. During evaluation, this conservative step-limit of $7500$ steps is removed and instead a maximum of five minutes running time is used.

\item[$\mathcal{D}, I$:]
The target problems consist of $100$ training and $100$ disjoint test problem instances taken from each of six different domains from the international planning competition (IPC).
The instances, however, were not taken from a particular round of the IPC as some domains only contain few instances. Instead, \shortciteA{speck-icaps21} used instance generators to generate instances that resemble those of the IPC tracks.

\item[$\Pi$:]
The policies are constrained to be a function of a specified observable state. 
The observable state consists of simple statistics about the heuristics in the configuration space.
Specifically, for every heuristic $h$, it contains the
\begin{inparaenum}[(i)]
\item maximum $h$ value;
\item minimum $h$ value;
\item average $h$ value;
\item variance of $h$ over all possible next states;
\item number of possible next states as determined by $h$; and
\item current expansion step $t$.
\end{inparaenum}
In order to encode progress towards solving a problem instance, \shortciteA{speck-icaps21} did not use these state features as is, but rather their change w.r.t. the previous step (i.e., the difference between consecutive observations).

\item[$c$:]
The considered cost metric is the total number of node expansions, i.e., the number of planning steps until a solution is found. The decomposed cost metric is $+1$ for every step. Thus, configuration policies that minimize the average number of planning steps are preferential. Note that, given the termination criterion of $\mathcal{A}$, the maximal cost at configuration time is $7500$, corresponding to not finding a solution in time. During evaluation, coverage is analyzed instead, i.e., the number of instances solved within the five minute budget.
\end{description}

\paragraph{Solution Method:}
The proposed solution approach by \shortciteA{speck-icaps21} uses a small double deep Q-network~\shortcite<\text{DDQN},  >{van-hasselt-aaai16} to learn a dynamic configuration policy via reinforcement learning. In our experiments, we use the original reinforcement learning code with the exact same hyperparameters as provided by the original authors. Since DACBench offers a standard RL interface (see Section~\ref{sec:dacbench}), the original RL code could be reused without modification.

\paragraph{Experimental Setup:}
In our experiments, we make use of the implementation of the interface as provided via DACBench (FastDownward benchmark). Following \shortciteA{speck-icaps21}, we learn a separate policy for each domain, however, to reduce the computational cost, we limit ourselves to a representative set of three out of six domains. 
Following \shortciteA{speck-icaps21}, we perform five independent training runs for each domain. In each training run, an RL agent experiences $10^6$ steps of the planning system, taking 8-12 hours on our system.  Since $|\Theta| = 4$, $\sbs$ and $\vbs$ could be determined exactly for each domain by evaluating $c(\vect{\theta},i)$ for all (4 $\times$ 100) combinations. For Hydra, we used a maximum portfolio size of three which is sufficient to cover the optimal portfolio.

\begin{figure}
\begin{minipage}{.5\linewidth}
\centering
\subfloat[\textsc{barman} domain]{\label{main:a}\includegraphics[width=\textwidth]{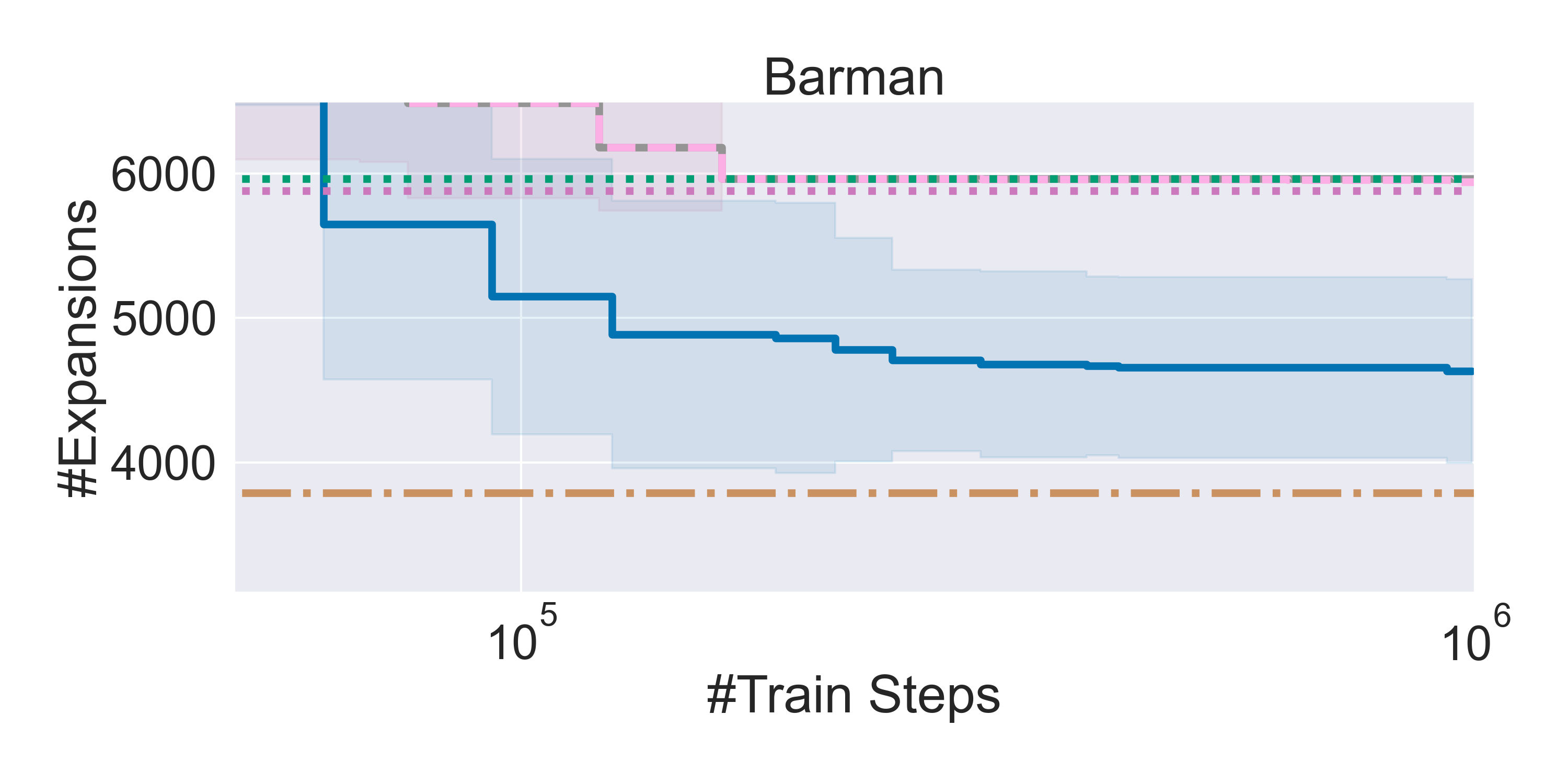}}
\end{minipage}%
\begin{minipage}{.5\linewidth}
\centering
\subfloat[\textsc{blocksworld} domain]{\label{main:b}\includegraphics[width=\textwidth]{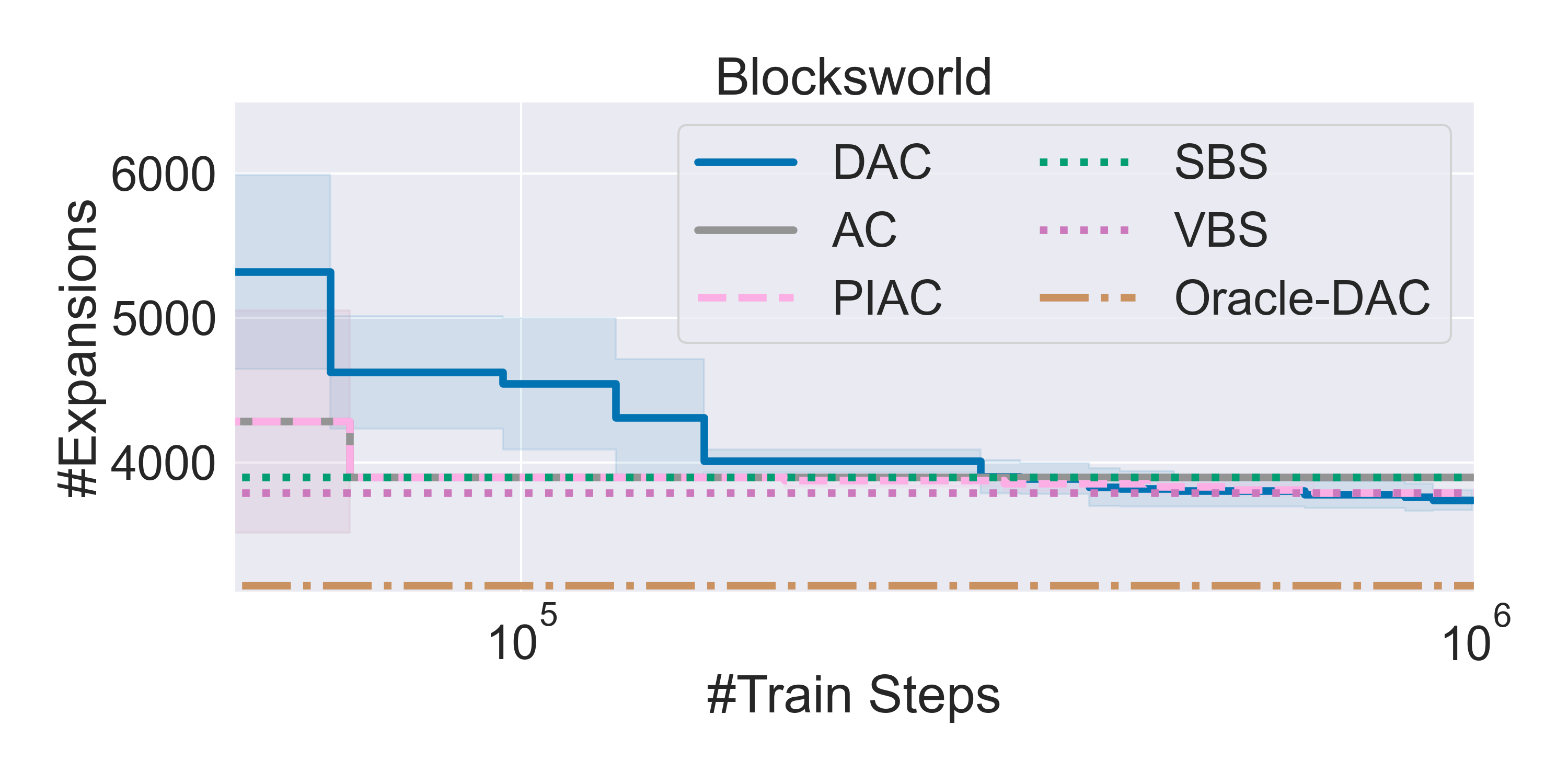}}
\end{minipage}
\centering
\subfloat[\textsc{visitall} domain]{\label{main:c}\includegraphics[width=.5\textwidth]{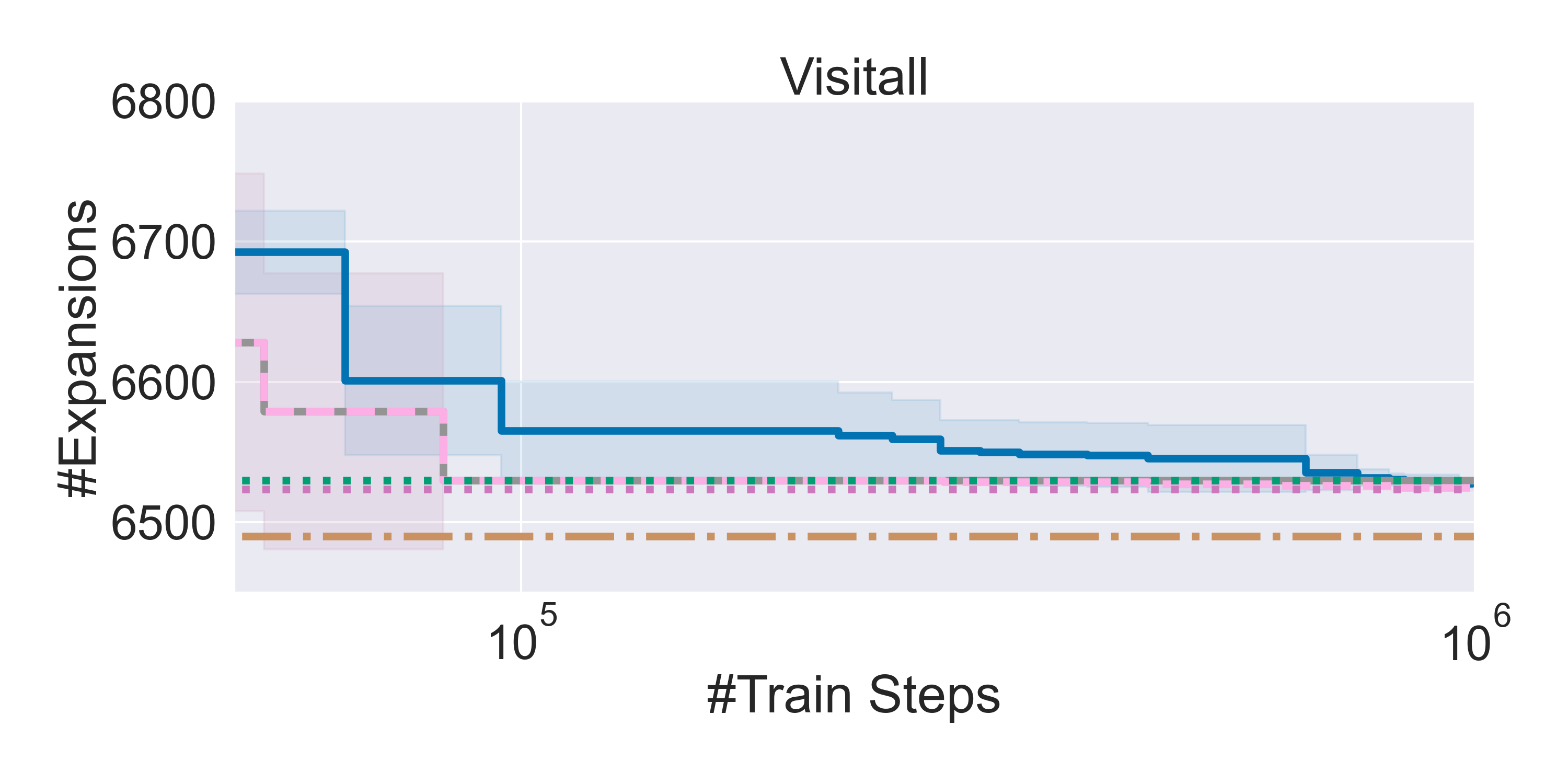}}
\caption{Incumbent performance of \dac{DDQN}, \hydra{}, and \smac{} when determining a heuristic selection policy for FastDownward on (a) the \textsc{barman}, (b)~\textsc{blocksworld}, and (c)~\textsc{visitall} domains. Solid lines depict the mean of five independent configuration runs and the shaded area the standard deviation.
SBS depicts the single best configuration and VBS the oracle configuration portfolio across all instances. Oracle-DAC is the oracle portfolio of all dynamic policies evaluated by \dac{DDQN}, providing a pessimistic performance estimate of optimal dynamic configuration policy.}
\label{fig:fd_dac_vs_ac}
\end{figure}

\paragraph{Results:}
Figure~\ref{fig:fd_dac_vs_ac} compares the anytime performance of \dac{DDQN} to that of \smac{} and \hydra{} for all three domains.
On the \textsc{barman} domain, \dac{} finds policies that on average clearly outperform the best static baseline in less than 10\% of the total budget.
On the \textsc{blocksworld} domain, \dac{} almost needs the full budget, but eventually finds policies that marginally outperform the $\vbs$.
The \textsc{visitall} domain is even slightly harder and \dac{DDQN} does not confidently find policies outperforming the static baselines within the limited budget.
For lower budgets, \ac{} and \piac{} obtain clearly better policies on \textsc{visitall} / \textsc{blocksworld}, and \piac{} eventually approaches $\vbs$ performance on both domains.
Table~\ref{tab:ai_planning_test_coverage} compares the coverage results for all learned policies on the test problem instances with a static baseline. Here, we find that our learned policies generalize well to the test scenarios and achieve similar coverage as reported in the original paper. In the \textsc{barman} domain, DAC policies dominate, and while we achieve a slightly lower coverage on this domain than originally, this can largely be attributed to an individual training run of ours performing worse than the others, with the individual coverages $85.00$, $88.32$, $67.00$, $84.00$, and $84.00$. In the other two domains, we achieve slightly higher coverages than originally, and the DAC policies perform similarly well as the best static policies.
\begin{table}
    \centering
    \resizebox{1\textwidth}{!}{%
    \begin{tabular}{l rrrrr r rrrr c}
        \toprule
         Algorithm & 
         \multicolumn{6}{c}{\textsc{dac policy}} & 
         \multicolumn{4}{c}{\textsc{single heuristic}} & 
         \multicolumn{1}{c}{\textsc{as oracle}} \\%
         \cmidrule(lr){2-7} \cmidrule(lr){8-11} \cmidrule(lr){12-12} %
         \multirow{2}{*}{Domain (\# Inst.)} & 
         \multicolumn{5}{c}{\textsc{rl}} & 
         \multirow{2}{*}{\textsc{rl$^\dag$}} & %
         \multirow{2}{*}{$h_{\text{ff}}$} & 
         \multirow{2}{*}{$h_{\text{cg}}$} & 
         \multirow{2}{*}{$h_{\text{cea}}$} & 
         \multirow{2}{*}{$h_{\text{add}}$} & %
         \multirow{2}{*}{\textsc{single $h$}} \\%
         {} & \tiny{Run\#1} & \tiny{Run\#2} & \tiny{Run\#3} & \tiny{Run\#4} & \tiny{Run\#5} & {} & {} & {} & {} & {} & {}\\%
         \cmidrule(lr){1-1} \cmidrule(lr){2-7} \cmidrule(lr){8-11} \cmidrule(lr){12-12}%
         \rowcolor{Gray}
         {}      & \multicolumn{5}{c}{81.7} &  & & & & & \\
         \rowcolor{Gray}
         \multirow{-2}{*}{\textsc{barman} (100)} & \tiny{85.0} & \tiny{88.3} & \tiny{67.0} & \tiny{84.0} & \tiny{84.0} & \multirow{-2}{*}{84.4} & \multirow{-2}{*}{66.0} & \multirow{-2}{*}{17.0} & \multirow{-2}{*}{18.0} & \multirow{-2}{*}{18.0} & \multirow{-2}{*}{67.0}  \\
         
         \multirow{2}{*}{\textsc{blocksworld} (100)} & \multicolumn{5}{c}{93.6} & \multirow{2}{*}{92.9} & \multirow{2}{*}{75.0} & \multirow{2}{*}{60.0} & \multirow{2}{*}{92.0} & \multirow{2}{*}{92.0} & \multirow{2}{*}{93.0} \\
         {} & \tiny{95.0} & \tiny{95.0} & \tiny{91.0} & \tiny{94.0} & \tiny{93.0} & & & & & & \\
         
         \rowcolor{Gray}
         {}   & \multicolumn{5}{c}{58.6} &  & & & & &  \\
         \rowcolor{Gray}
         \multirow{-2}{*}{\textsc{visitall} (100)}  & \tiny{58.1} & \tiny{56.1} & \tiny{57.8} & \tiny{60.0} & \tiny{61.0} & \multirow{-2}{*}{56.9} & \multirow{-2}{*}{37.0} & \multirow{-2}{*}{60.0} & \multirow{-2}{*}{60.0} & \multirow{-2}{*}{60.0} & \multirow{-2}{*}{60.0}\\
         
         \cmidrule(lr){1-1} \cmidrule(lr){2-7} \cmidrule(lr){8-11} \cmidrule(lr){12-12}%
         \textsc{sum} (300) & \multicolumn{5}{c}{233.9} & 234.2 & 178.0 & 137.0 & 170.0 & 170.0 &  220.0 \\
         \bottomrule
    \end{tabular}%
    }
    \caption{Number of solved \emph{unseen} test problem instances averaged over five independently repeated training runs. Column \textsc{rl} provides the results of our experiment, with the results of the individual runs given in a smaller font, whereas \textsc{rl$^\dag$} contains the original coverage values as reported by \shortciteA{speck-icaps21}. All $h_i$ columns contain the number of solved problem instances when only using the specific heuristic. \textsc{as oracle} reports the coverage results of the theoretically best algorithm selector.}
    \label{tab:ai_planning_test_coverage}
\end{table}
\paragraph{Discussion:}
Our results confirm the results of \shortciteA{speck-icaps21} where the DAC policies (i) obtain slightly lower coverage than the single best heuristic in the \textsc{visitall} domain, (ii) outperform the single best heuristic and are close in performance to the theoretical best algorithm selector on the \textsc{blocksworld} domain and (iii) provide the best coverage by far in the \textsc{barman} domain.
Most notably, on average, the learned DAC policies are capable of solving more problem instances than the theoretical best algorithm selector, which already provides a significant improvement over using the single best heuristic.
Our analysis of the approach's anytime performance also revealed that when less time is available, static AC approaches, in particular \hydra, outperform \dac{DDQN} on two of the three domains. However, on the remaining domain (\textsc{barman}), superior dynamic policies are easily found. It is worth noting that on all three domains, oracle-DAC is clearly superior, suggesting the potential to further improve performance by using better DAC methods and/or more informative state features.

\subsection{Learning Rate Control in Neural Network Training}
\label{sec:sgd}
\shortciteA{daniel-aaai16} investigated meta-learning a controller for the learning rate hyperparameter $\eta$ in Stochastic Gradient Descent (SGD) style neural network optimizers.
SGD is the method of choice for optimizing the parameters $\vect{w}$ of 
deep neural networks, i.e., solve $\argmin_{\vect{w}} L(\vect{w}, D)$, where $L$ is some differentiable measure of loss on the training data $D$. In deep learning, it is common to have millions of parameters. To scale up to such extremely  high-dimensional $\vect{w}$, SGD exploits the fact that $\nabla_{\vect{w}} L(\vect{w}, D) =  \frac{\partial L(\vect{w}, D)}{\partial \vect{w}}$ can be computed exactly, and updates $\vect{w}$ in the opposite direction of the gradient.
As datasets in deep learning are huge, 
computing the ``full batch'' gradient is typically too expensive. Instead, SGD computes the gradient at every optimization step for a different randomly selected ``mini-batch'' $B \subset D$.
While this gradient is an unbiased estimate of the actual gradient, i.e., $\mathbb{E}[\nabla_{\vect{w}} L(\vect{w}, B)] = \nabla_{\vect{w}} L(w, D)$,
variance can cause gradients to occasionally point in the wrong direction.
Furthermore, gradients only provide local information and do not tell us how far we can move without overshooting. Moreover, the optimal step sizes per dimension may vary strongly, a problem known as ill-conditioning. Over the last decade a variety of different variants of SGD, e.g., Momentum~\shortcite{jacobs-nn88a}, RMSprop~\shortcite{tieleman-lecture12}, and Adam~\shortcite{kingma-arxiv14a}, have been proposed that aim to address these and other issues. However, despite their popularity, modern SGD variants are still sensitive to their hyperparameter settings. In particular, they still have a global/initial learning rate $\eta$, that uniformly scales the step taken in each dimension, and that must typically be optimized for the problem at hand~\shortcite{bengio-bookchapter12a}. When setting $\eta$ too low, optimization is slow, while too high $\eta$ might even lead to divergence. 
To the best of our knowledge, \shortciteA{daniel-aaai16} was the first work that explored replacing $\eta$ by a meta-learned controller, producing more robust SGD methods. \shortciteA{xu-arxiv19-apalike} followed up on this idea, and most recently~\shortciteA{almeida-arxiv21} considered meta-learning the control of learning rate \emph{and} various other hyperparameters (e.g., weight-decay and gradient clipping).

\paragraph{Problem Formulation:}
The meta-learning approach by \shortciteA{daniel-aaai16} is readily seen as automated DAC. Below, we briefly detail each of the DAC components:
\begin{description}
\item[$\mathcal{A}$, $\Theta$:] \shortciteA{daniel-aaai16} present a general method for dynamically configuring the learning rate $\eta_t \in \mathbb{R}^+$ at every optimization step of SGD. In their experiments, they do this for two SGD variants: RMSprop and Momentum.\footnote{In the original paper, Momentum was simply referred to as ``SGD''.} Note that in the first optimization step, a fixed learning rate $\eta_0$ is used.
\item[$\mathcal{D}, I$:] Instances correspond to neural network optimization problems, and are represented by the quadruple $\langle D, L, k, \xi \rangle$, where
\begin{itemize}
    \item $D$ is the data we want to fit the neural network to. In their experiments, \shortciteA{daniel-aaai16} consider image classification, using examples from the MNIST and CIFAR10 datasets.
    \item $L$ is the differentiable loss function to be minimized. Daniel et al.\ used
    cross-entropy loss, i.e., the negative log-likelihood of the data $D$ under the model with parameters $w$, where this model can be any parametric model.
    \shortciteA{daniel-aaai16} used small convolutional neural networks (CNNs).
    \item $k$ is the cutoff: SGD is terminated after $k$ optimization steps.
    \item $\xi$ is the seed of the pseudo-random number generator used for random neural network initialisation and mini-batch sampling.
\end{itemize}
\item[$\Pi$]: \shortciteA{daniel-aaai16} considered dynamic configuration policies that are a log-linear function $\pi_{\vect{\lambda}}(\vect{\phi}) = \exp(\lambda_0 + \sum_{j=1}^4 \lambda_j \phi_j)$ of four expert features $\vect{\phi}$ that in turn depend on the previous learning rate $\eta_{t-1}$ and the current loss/gradients for each data point. See the original paper for a detailed description of $\vect{\phi}$.
\item[$c$]: \shortciteA{daniel-aaai16} aim to control the learning rate $\eta$ as to maximally reduce the training loss. Specifically, the cost of a run is quantified as $\min(\frac{1}{k-1}\log(\frac{E_{k}}{E_{1}}), 0)$,
where $E_{t}$ is the full batch training loss after $t$ optimization steps. Note that we handle divergence cases by setting the costs of runs that fail to reduce the training loss to 0.

\end{description}


\paragraph{Solution Method:}
\shortciteA{daniel-aaai16} solved this DAC problem by directly optimizing the policy parameters $\vect{\lambda}$ using the Relative Entropy Policy Search (REPS) policy gradient method. In our experiments, we will also optimize $\vect{\lambda}$ directly, but instead use Sequential Model-based Algorithm Configuration  \shortcite<SMAC,>{hutter-lion11a}. Note that we follow a \emph{DAC by static AC}, instead of a \emph{DAC by RL} approach (see Section~\ref{sec:opt4dac}).
This decision was motivated by the fact that \shortciteA{daniel-aaai16} provide too little details about the method and its implementation, to allow us to confidently reproduce the original meta-training pipeline.
On the other hand, SMAC is a popular open source~\shortcite{lindauer-jmlr22a} tool for Bayesian optimization that we conjecture to be suitable to reliably and globally optimize $\vect{\lambda}$ within a reasonable time frame.

\begin{table}[tbp]
\scriptsize
    \centering
    \resizebox{1\textwidth}{!}{%
    \begin{tabular}{lccccc}
        \toprule
        & $\mathcal{A}$ & $D$ & $L$ & k & $\xi$ \\
        \hline
         meta-training & RMSprop & MNIST-small& c-p-c-p-c-r-fc-s (varying \# filters) & 300-1000 & varying\\
         & Momentum & MNIST-small& c-p-c-p-c-r-fc-s (varying \# filters) & 300-1000 & varying\\
         \midrule
         meta-testing & RMSprop & MNIST  & c-p-c-p-c-r-fc-s (20-50-200 filters) & 2000 & fixed\\
         & Momentum & MNIST  &  c-p-c-p-c-r-fc-s (20-50-200 filters) & 2000 & fixed\\
         & RMSprop & CIFAR10  & c-p-r-c-r-p-c-r-p-c-r-fc-s (32-32-64-64 filters)  & 6000  & fixed\\
         & Momentum & CIFAR10  & c-p-r-c-r-p-c-r-p-c-r-fc-s (32-32-64-64 filters) & 12000 & fixed\\
         \bottomrule
    \end{tabular}%
    }
    \caption{A summary of the six different DAC setups used in \shortcite{daniel-aaai16}. During meta-training, 100 target problem instances are considered, generated by randomly varying $D$ (dataset), $L$ (loss), $k$ (cutoff), and $\xi$ (seed). The meta-testing setups consider a single instance. MNIST-small: To avoid bias towards specific training examples, a randomly varied subset of 6K-30K of the 60K MNIST training examples is used during meta-training. The losses $L$ differ only in the predictive model. All use CNNs, but a different layered architecture (c: same convolution with 3x3 filter, r: ReLU, fc: fully connected, s: softmax). To avoid bias towards specific architectures, the number of filters used is varied randomly in meta-training (in ranges [2-10]-[5-25]-[50-250]).}
    \label{tab:sgdsetups}
\end{table}

\paragraph{Experimental Setup:}
In our experiments, we used the DACBench implementation of the DAC scenario described above (SGD-DL). Apart from using SMAC instead of REPS, we aimed to maximally replicate the setup used in the original paper. Note that \shortciteA{daniel-aaai16} actually considered six slightly different scenarios: Two for learning the $\eta$-controller for RMSprop/Momentum, resp., and two for testing each of the meta-learned controllers on MNIST/CIFAR10, resp. The differences between these setups are summarized in Table~\ref{tab:sgdsetups}. As we did not have access to the original code, replication was restricted by the details disclosed in the original paper.\footnote{We also contacted Chris Daniel, the first author, but he did not have access to the proprietary code anymore, either, and was thus not able to help us replicate the original setup.} The remaining design choices were mostly made heuristically. Some had to be optimized to obtain similar baseline behavior. Here, we found the use of a sufficiently large mini-batch size (64 at meta-training, 512 at meta-testing), and Xavier weight initialisation, to be particularly important. For meta-training the two $\eta$-controllers, we used a meta-training set $I' \sim \mathcal{D}$ of 100 instances and the default parameter settings of SMAC, and optimized $\vect{\lambda} \in [-10, 10]^5$, using a symmetric log-scale with linear threshold $10^{-6}$, for 5000 inner training runs. Each SMAC run took less than 2 CPU-days on our system. To assess meta-training variability, we performed five such runs in parallel, selecting the configuration with the highest meta-training performance for meta-testing. For Hydra, we used the same parameters as SMAC, and a maximum portfolio size of 10. Finally, to determine $\sbs$ and $\vbs$, we discretized $\Theta$ (1000 values, log-scale in [$10^{-5}, 10^{0}$]) and evaluated $c(\vect{\theta},i)$ for all (1000 $\times$ 100) combinations.

\begin{figure}%
    \centering
    \subfloat[\centering Meta-Training RMSprop]{{\includegraphics[width=0.5\textwidth]{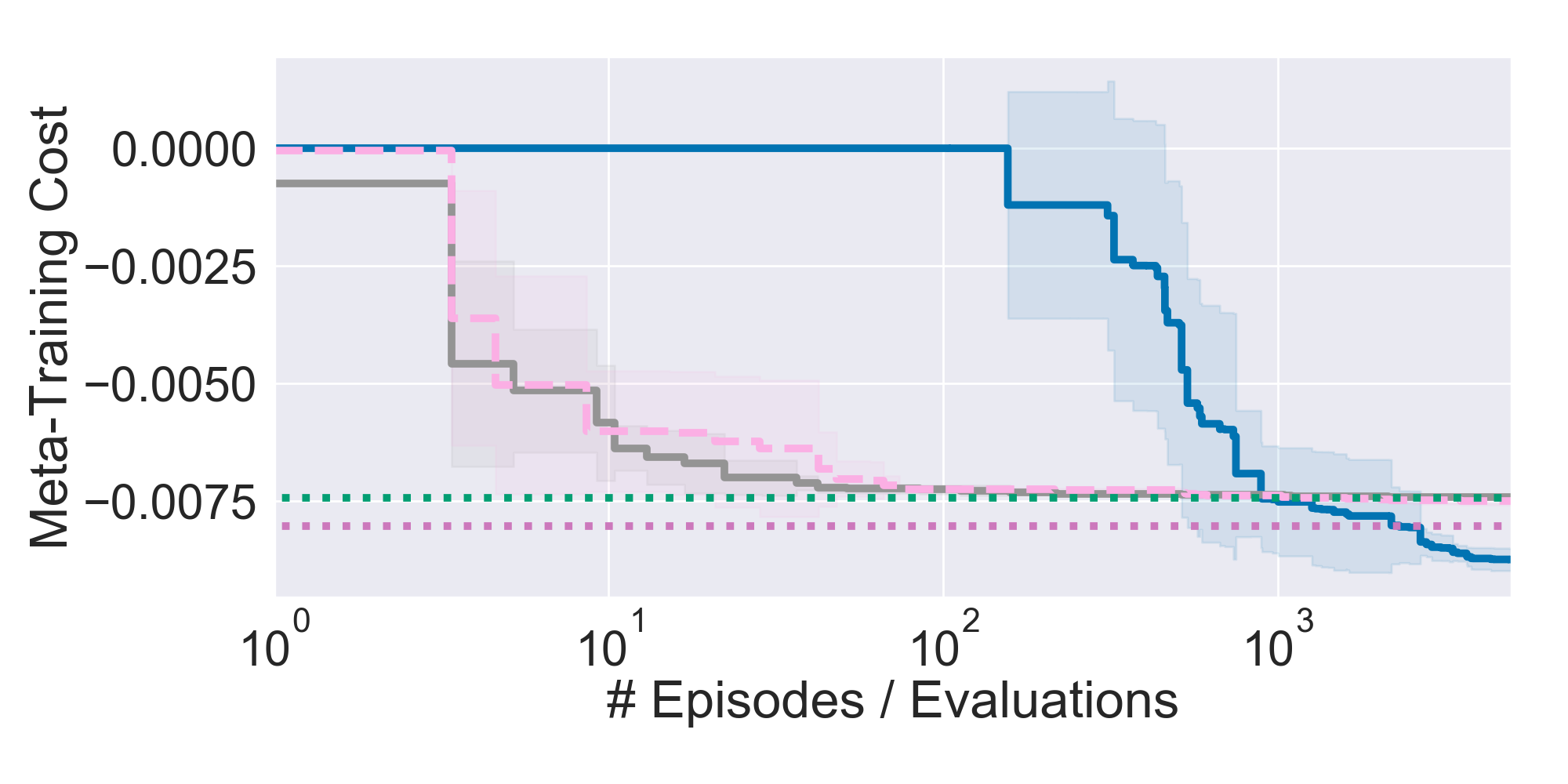} }}%
    \subfloat[\centering Meta-Training Momentum]{{\includegraphics[width=0.5\textwidth]{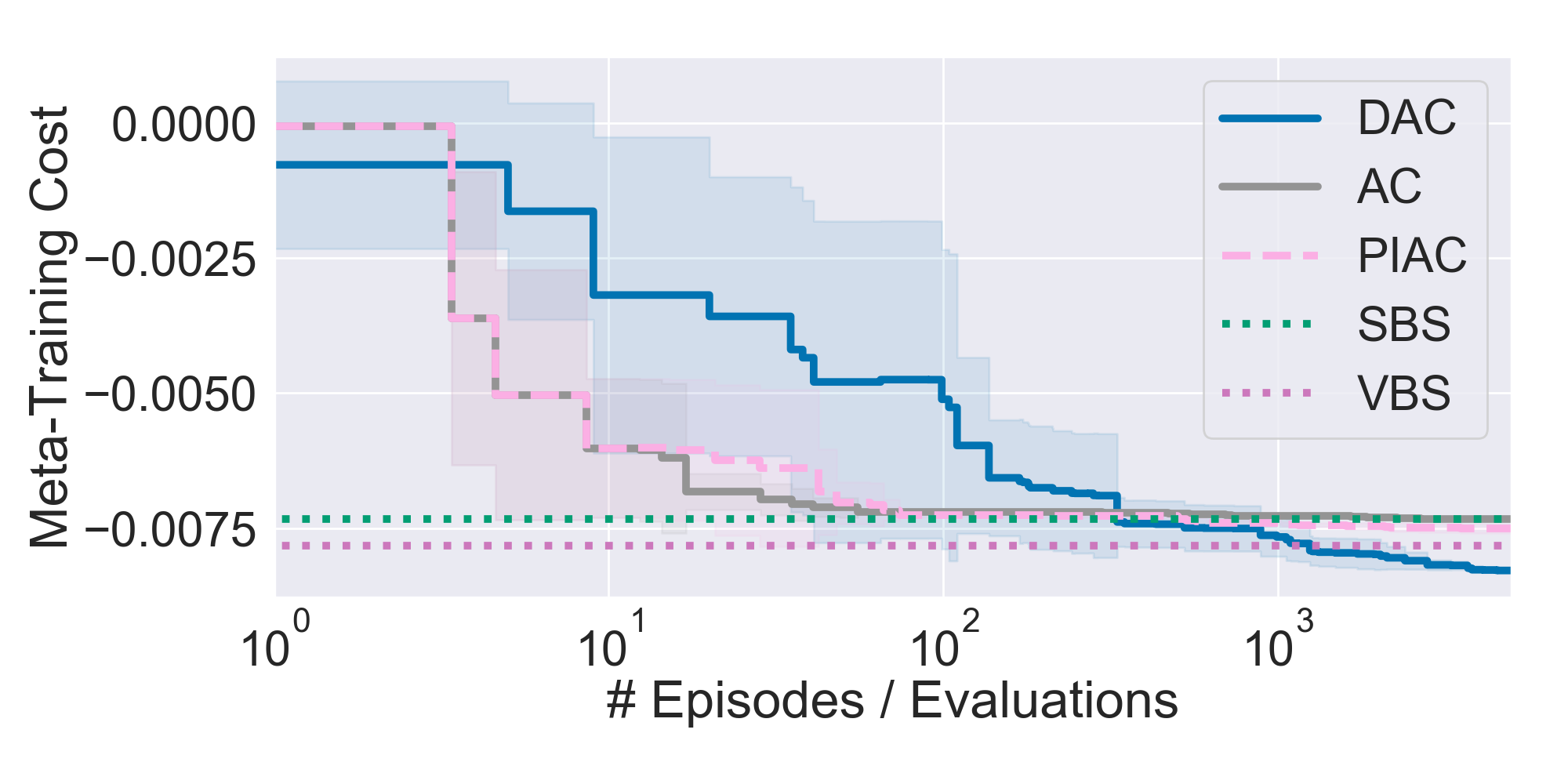}}}%
    \caption{Incumbent performance of \dac{SMAC configuring a parametric DAC policy}, \piac{}, and \ac{} when meta-learning learning rate configuration for RMSprop (left) and Momentum (right). Solid lines depict the mean of five independent meta-learning runs and the shaded area the standard deviation. $\sbs$ depicts the single best configuration and $\vbs$ the oracle configuration selection portfolio across all instances.}%
    \label{fig:sgd_dac_vs_ac}%
\end{figure}

\begin{figure}
    \centering
    \includegraphics[width=\textwidth]{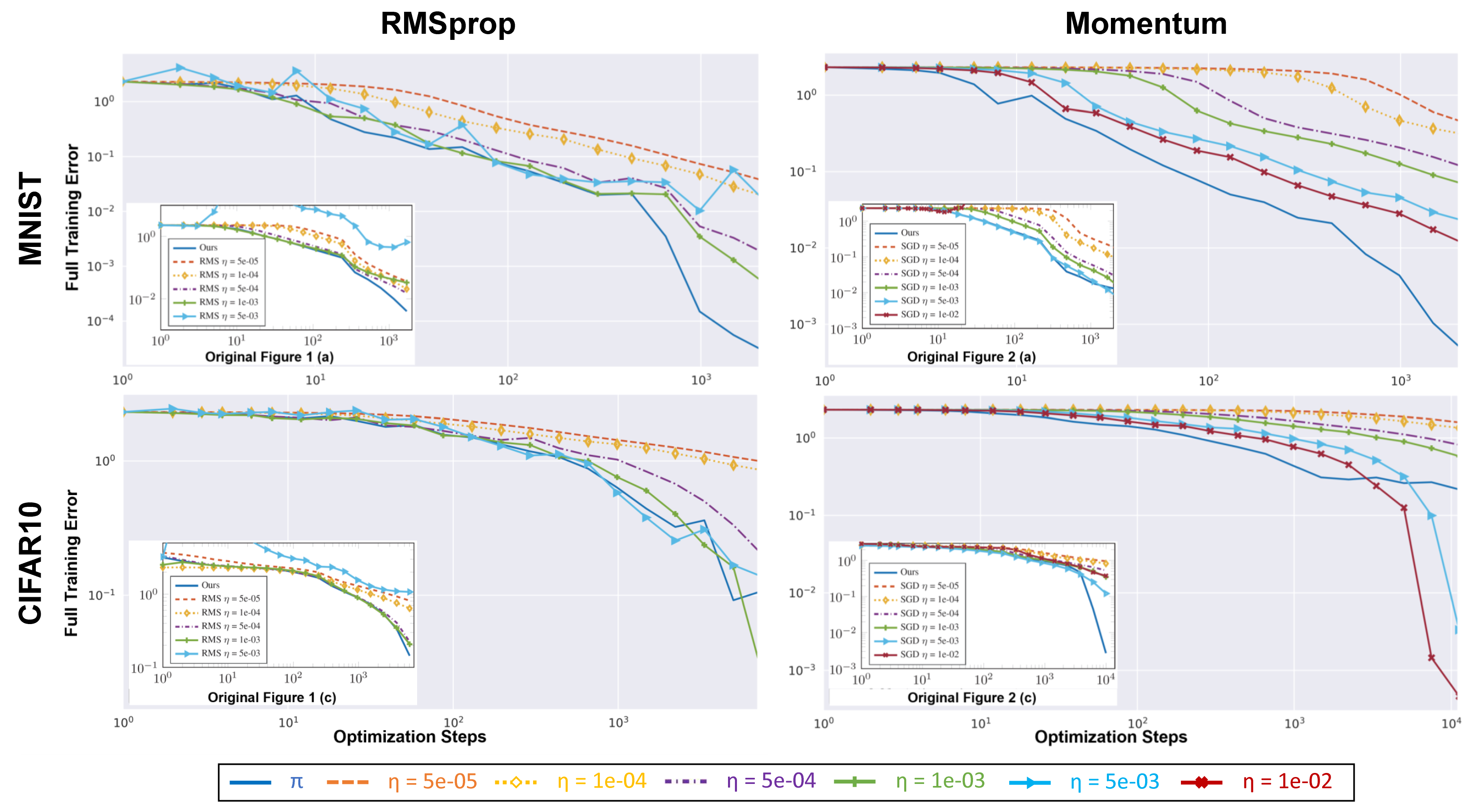}
    \caption{Comparison of learning curves for RMSprop/Momentum using the meta-learned $\eta$-controllers ($\pi$) vs. several static baselines, on MNIST/CIFAR10. Each dataset/optimizer combination appears in its own sub-figure. For ease of comparison, the corresponding figure in the original paper is shown in the bottom left corner of each sub-figure.\label{fig:cd}}
\end{figure}

\paragraph{Results:}
Figure~\ref{fig:sgd_dac_vs_ac} compares the anytime performance of \dacsmac{} to that of \hydra{} and \smac{} for RMSprop (left) and Momentum (right).
In both cases, \dac{}'s initial performance is worse than its static counterparts.
This difference in relative performance is most blatant for RMSprop, where \dac{} takes over $100$ evaluations to find a non-diverging policy (i.e., with negative average cost), while \ac{} achieves near $\sbs$ performance in that time.
\hydra{} only marginally improves upon \smac{} and $\sbs$, and does not attain $\vbs$ performance.
Despite the slow start, all \dac{} runs eventually outperform all \ac{} and \piac{} runs, ultimately attaining a policy that reduces training loss 0.71\% (RMSprop) and 0.46\% (Momentum) more per step than the VBS on average ($\sim$ 58\% and 35\% after 650 steps).
Figure \ref{fig:cd} shows the full batch training loss $L(w_t, D)$ at each optimization step using the meta-learned $\eta$-controller that performed best in meta-training, and various static baselines, in each of the four meta-test setups.
Overall, the training curves for our baselines look similar to the original, both in terms of absolute and relative performance. An exception are high learning rates. For RMSprop, our curves look quite different, but are similarly chaotic. For momentum, the highest learning rate performs best for us, while the original diverges.
On MNIST, both meta-learned controllers ($\pi$) clearly outperform the best static baseline, even though the cutoff $k$ is two times higher than the highest cutoff considered during meta-training. This result is similar to that of the original paper, but our learned controller arguably even does better.
On CIFAR10, the meta-learned controller ($\pi$) performs similar to (RMSprop), or better than (Momentum) the best baseline in the first 1000 update steps, but fails to achieve the best final performance. Here, our results differ from the original, where the final performance was similar (RMSprop) or better (Momentum) than the best static baseline.

\paragraph{Discussion:} The ``flavour of AC'' prevailing on these scenarios depends on the budget available: For sufficiently large budgets ($>$ 1000 evaluations), \dac{} confidently outperforms static AC. However, \dac{} is clearly outperformed by \ac{} for smaller budgets.
While the best DAC policies are better, arbitrary static policies tend to outperform arbitrary DAC policies for this scenario, e.g., the vast majority of DAC policies diverge for RMSprop. Nonetheless, the poor relative initial performance of \dac{} is not inherent, and could, e.g., be addressed by using a different initial design that prioritizes static policies (${\pi_{\vect{\lambda}}:\lambda_k = 0, \forall \, k > 0}$). Also note that we cannot compare our meta-training results to those obtained by REPS, since \shortciteA{daniel-aaai16} did not analyze meta-training. Our meta-testing results, however, validate that the SGD-DL benchmark considers a highly similar setup and that it can be used to learn controllers that perform similarly well as in the original paper. On the other hand, we also observed differences that are unlikely explained by random noise alone. For example, momentum seems to prefer higher learning rates in our experiments, and our meta-learned controller does not transfer as well to higher cutoffs on CIFAR10. Finally, the configurations $\vect{\lambda}$ we found differ strongly from those reported in the original paper, and using the latter even caused divergence in our experiments. We currently cannot explain those differences, and lacking the original code, further insight can only be gained through trial \& error. We emphasize that, in contrast to the original code, our benchmark is publicly available to facilitate future research on DAC. 

\section{Conclusion}
\label{sec:conclusion}

To conclude, we again summarize our main insights and results, and discuss possible future research directions opened up by this work.

\subsection{Summary}
\label{sec:summary}
In this article, we presented the first comprehensive overview of automated Dynamic Algorithm Configuration (DAC), a novel meta-algorithmic framework proposed by \shortciteA{biedenkapp-ecai20a}. To this end, we introduced automated DAC as a natural extension of previous research efforts in automated static algorithm configuration and manual DAC. Furthermore, we situated automated DAC in a broader context of AI, discussing how it can be viewed as a form of ``semi-automated'' programming, as a generalization of existing meta-algorithmic frameworks, and as an automated approach to the design of operator selection and parameter control mechanisms. After formalizing DAC further, we introduced its methodology and showed how prior art can be roughly subdivided in two schools, tackling the problem using \emph{reinforcement learning} and \emph{optimization} methods, respectively. 
On the empirical side, we presented and extended DACBench, a novel benchmark library for DAC proposed by \shortciteA{eimer-ijcai21} and showed that DAC can be successfully applied to evolutionary optimization, AI planning, and machine learning. As the first paper, we provided thorough empirical evidence that automated DAC can outperform prior static AC methods. In summary, we found that on all scenarios considered, automated DAC discovered policies that were at least as good as, and typically better than, their static counterparts. Depending on the scenario, this sometimes required less, but usually more (up to 10$\times$) computational budget than a state-of-the-art static AC method needed to converge on the same scenario, on average.

\subsection{Limitations and Further Research}
\label{sec:future}
While these case studies and other previous  applications provide a ``proof of concept'' for automated DAC, we point out that much remains to be done to unlock its full potential, and we hope that this work may serve as a stepping stone for further exploring this promising line of research. In what remains, we will discuss some of the limitations of contemporary work and provide specific directions for future research.

\paragraph{Jointly configuring many parameters:}
While static approaches are capable of jointly configuring hundreds of parameters, the configuration space in contemporary DAC is typically much smaller, often considering only a single parameter. While the configuration space is smaller, the candidate solution space (i.e., the dynamic configuration policy space) grows exponentially with the number of reconfiguration points, in the worst case, and is thus typically drastically larger than static configuration policy spaces.
Although modern techniques from reinforcement learning scale much better than ever before, we still know too little about the internal structure of DAC problems to handle this exploding space of possible policies. For example, not much is known regarding interaction effects of parameters in the DAC setting. If there should be only a few interaction effects between parameters as in static AC~\shortcite{hutter-icml14a,wang-jair16a}, learning several independent policies might be a way forward. 

\paragraph{Temporally sparse DAC:}
Note that not all parameters can/must be reconfigured at every time step. Also, our results suggest that an initial bias towards static configuration policies could improve the anytime performance of DAC in various scenarios. Mixed static and dynamic configuration, and learning ``when to reconfigure''~\shortcite{biedenkapp-icml21} therefore present one opportunity to scaling up DAC. Furthermore, we plan to extend the DAC formalism with \emph{partial} reconfiguration to capture intrinsic temporal conditionalities, e.g., a parameter not being used in some execution steps.

\paragraph{Warm-starting DAC:} 
Most prior art derives dynamic configuration policies from scratch, while in many cases good default parameter control mechanisms are known. Beyond strong baselines, these existing policies could also be used to warm-start the automated process. This idea has already been explored by \shortciteA{shala-ppsn20a} (see also Section~\ref{sec:cmaes}), but could be extended in various ways. For example, we could learn from an ensemble of teachers to exploit their complementary strengths.

\paragraph{Online DAC:} 
Most prior art performs algorithm configuration offline, i.e., the optimal static/dynamic configuration policy is derived in a dedicated configuration phase proceeding use/test time (see Figure~\ref{fig:offline_v_online}).\footnote{As discussed in Section~\ref{sec:manual-dac}, we do not regard the majority of previous ``online learning approaches'' to parameter control as prior art in \emph{automating} DAC.}
However, when using the target algorithm, more information about the target problem distribution and relative performance of candidate policies becomes available, and \emph{online algorithm configuration} approaches \shortcite{fitzgerald-phd21} capable of exploiting this information and transferring experience across test instances,  continually refining the policy are an interesting direction of future research.

\paragraph{Better DAC methods:}
Successful DAC requires more than just computational resources. To apply DAC, a practitioner must make many choices that critically affect not just its efficiency, but also its effectiveness. As a consequence, key ingredients for successful DAC are currently (i) target domain expertise, (ii) DAC methodology expertise, and (iii) trial and error. Note that this conflicts with the main objective of automated DAC, i.e., reducing reliance on human effort and expertise. To address this shortcoming, we need better methods. In particular, as discussed in Section~\ref{sec:rl_vs_opt}, we believe that there is a need for dedicated dynamic algorithm configuration packages capable of combining the strengths of the contemporary DAC by reinforcement learning and optimization approaches.

\paragraph{Domain-Expert driven DAC:}
From working on static AC for more than a decade, we know that a challenge AC poses to users is to specify the inputs, including questions such as: (i)~Which instances will reflect future real-world use cases well? (ii)~Which parameters are important and should be configured, and using which domain (upper \& lower bounds, etc)? (iii)~Which metric will accurately quantify the true desirability of a configuration or policy? This hinders the adoption of such meta-algorithmic approaches in practice. To tackle this problem, we envision a new paradigm which is driven by the domain expert and allows for monitoring of the training and deployment performance, and for live adjustments of training distributions, configuration spaces and performance metrics by the domain expert. Likewise, we would like to enable experts to express priors over the policies they would expect to work well, extending similar work in static AC~\shortcite{hvarfner2022pibo}. Finally, we would like domain experts to not only be able to steer DAC, but to also gain new and deeper insights from the automated search process, similar to various existing methods that capture the importance of hyperparameters in static AC~\shortcite{hutter-icml14a,biedenkapp-aaai17a,biedenkapp-lion18a,rijn-kdd18a,probst-jmlr19a}.

\paragraph{Extending DAC Benchmarks:} To stay relevant, these future directions will also have to be reflected in a set of contemporary DAC benchmarks, such as in DACBench, alongside continuing work on further expanding the scope of existing benchmarks. While DACBench covers a range of domains, some like SAT or MIP, which are commonly used in AC, are absent at the moment. Partnering with domain experts could help broaden the scope of DACBench and thus DAC in general. Beyond real-world benchmarks, there is also a need for additional ``toy'' benchmarks that permit efficient evaluation of DAC methods, something especially crucial in (i) the early stages of developing new methods and (ii) enabling meta-algoritmics to be applied to DAC itself. Finally, prior art, our own work included, rarely compares different DAC methods. To facilitate this, we need more than just benchmarks, we need a library of DAC methods and standard protocols to compare them.

\ifarxiv
\acks{
All authors acknowledge funding by the Robert Bosch GmbH.
Theresa Eimer and Marius Lindauer acknowledge funding by the German Research Foundation (DFG) under LI 2801/4-1. 
We thank Maximilian Reimer, Rishan Senanayake, G\"{o}ktu\u{g} Karaka\c{s}lı, Nguyen Dang, Diederick Vermetten, Jacob de Nobel and Carolin Benjamins for their contributions to DACBench, and Carola Doerr for the many discussions on related work and problem formulation.
}
\else
\fi

\clearpage

\bibliography{bib/strings,bib/local,bib/lib,bib/proc}
\bibliographystyle{theapa}

\clearpage
\appendix
\section{Problem-Theoretical Perspective on DAC}
\label{a:problemtheory}
In this Appendix, we present a theoretical motivation of why DAC (Definition~\ref{def:dac}) is a problem worth studying. In doing so, we provide grounding for many of the higher-level discussions in the main text. Since this kind of analysis is hardly standard, we start by introducing some fundamental concepts (Section~\ref{a:fundamentals}), then discuss the main results (Section~\ref{a:results}), and end with the formal justification (Section~\ref{a:proofs}).

\subsection{Fundamental Concepts}
\label{a:fundamentals}
\subsubsection{Computational Problems}
In this work, we formalized dynamic algorithm configuration (DAC) and related computational problems as follows:
\begin{definition}[Computational Problem \label{def:problem}]
In a computational problem $(\mathcal{X}, \mathcal{R})$, given any input $x \in X$, we are to compute an output $y$ satisfying $(x,y) \in \mathcal{R}$.
\end{definition}
Conceptually, each $(x, y) \in \mathcal{R}$ represents a problem instance $x$ and
a solution $y$ thereof. Note that instances may have more than one admissible solution, or even none at all. We will also use $R(x) = \{y|(x,y) \in R\}$ to denote the solution set for $x$. All problem definitions in this paper are structured syntactically as ``Given $x$ find a $y \in R(x)$''. 

\paragraph{Digression on Problem Classes:} Problems of this form are also known as ``search problems''. To avoid confusion, problem \emph{classes} group problems (e.g., DAC), not problem instances (e.g., DAC scenarios), i.e., when viewing DAC as a search problem (as in Definition~\ref{def:dac}), $\mathcal{X}$ would correspond to the set of all possible DAC scenarios and $\mathcal{R}(x)$ the set of optimal policies for some DAC scenario $x$.  The choice to restrict ourselves to search problems was a trade-off between 
(i) theoretical convenience / simplicity and (ii) expressiveness, i.e., alternative formulations exist that better model many of the problems we consider:
\begin{description}
\item[Optimization:] Can express that not every solution is equally good. For instance, in this work we use (i) ``find $x$ satisfying $x \in \argmax_x f(x)$'' rather than (ii) ``find $x$ maximizing $f(x)$''. Note that this difference, while subtle, is important for problem theory: For instance, if $\argmax_x f(x) = \emptyset$, in (i) we should return that no solution exists, while in (ii) we should return an as good as possible solution. 
\item[Distributional:] Can express that all inputs are \emph{not} equally likely, by modelling inputs as a distribution $\mathcal{D}$ rather than a set $\mathcal{X}$. Note that while we do not use distributional problems on the meta-level, we do use them as target problems.
\end{description}
However, for these problem classes, standard definitions for theoretical concepts such as \wordasword{reducibility} do not exist and any satisfactory definition would significantly complicate the reduction proofs in this appendix.

\subsubsection{Reducibility}
Problem formalization enables formal reasoning about the relationship between problems. In this appendix, we focus on a specific kind of relationship: Reducibility, as defined by \shortciteA[p. 506]{papadimitriou1994complexity}:
\begin{definition}[Many-one Reducibility ($m$-reducibility)]
\label{def:ms}
Let $(X,R)$ and $(X',R')$ be two computational problems, with $R: X \times Y$ and ${R': X' \times Y'}$. We say that $(X,R)$ is many-one reducible to $(X',R')$, or also $m$-reducible, which we denote ${(X,R) \leq_m (X',R')}$, if and only if computable functions
\begin{description}
\item[$\formulate$:] $X \rightarrow X'$
\item[$\interpret$:] $X \times \,Y' \rightarrow Y$
\end{description}
exist such that $\forall \ x \in X$ holds:
\begin{enumerate}
\item $(\formulate(x), y') \in R' \implies (x, \interpret(x,y')) \in R$
\item $R'(\formulate(x)) = \emptyset \implies R(x) = \emptyset$
\end{enumerate}
\end{definition}
Conceptually, (1) all solutions of the reformulated problem instance can be interpreted as a solution to the original problem instance, and (2) if the reformulated problem instance does not have any solutions, the original problem instance should neither.

When a problem $R$ is m-reducible to another $R'$, $R$ can be \emph{solved by reduction} to $R'$, i.e., given an algorithm $a'$ for $R'$, $a(x) = \interpret(x, a'(\formulate(x)))$ is an algorithm for $R$. It is worth noting that the existence of a many-one reduction does not necessarily render $R$ irrelevant: Solving $R$ by reduction to $R'$ may inherently increase computational \emph{complexity} since 
\begin{inparaenum}[(i)]
\item the reduction itself (i.e., $\formulate$, $\interpret$) may be costly
\item the reduction may abstract relevant info, making the reduced problem harder to solve.
\end{inparaenum}
The practical relevance of a reduction is further limited by the performance of known algorithms for $R'$.

Note that sometimes, even though one problem is not generally reducible to another, a special case is. We define this notion as
\begin{definition}[Conditional Reducibility]
Let $(X,R)$ and $(X',R')$ be two computational problems, with ${R: X \times Y}$ and ${R': X' \times Y'}$. We say that $(X,R)$ is \emph{conditionally} many-one reducible to $(X',R')$ under \emph{preconditions}~$c$, which we denote $(X,R) \leq^c_m (X',R')$, if and only if 
${(\{x \in X \, | \, c(x)\},R) \leq_m (X',R')}$, where $c$ is a Boolean function on $X$.
\end{definition}
The practical relevance of a conditional reduction additionally depends on how commonly its preconditions are satisfied. It is worth noting that many-one reducibility is \emph{transitive}, while conditional reducibility is not. However, the following holds: 
\begin{align*}
(X,R) \leq^c_m (X',R') \quad \land \quad (X',R') \leq_m (X'',R'') \quad \implies \quad (X,R) \leq^c_m (X'',R'')
\end{align*}

\subsection{Reducibility Results}
\label{a:results}
\begin{figure}
    \centering
    \includegraphics[width=\textwidth]{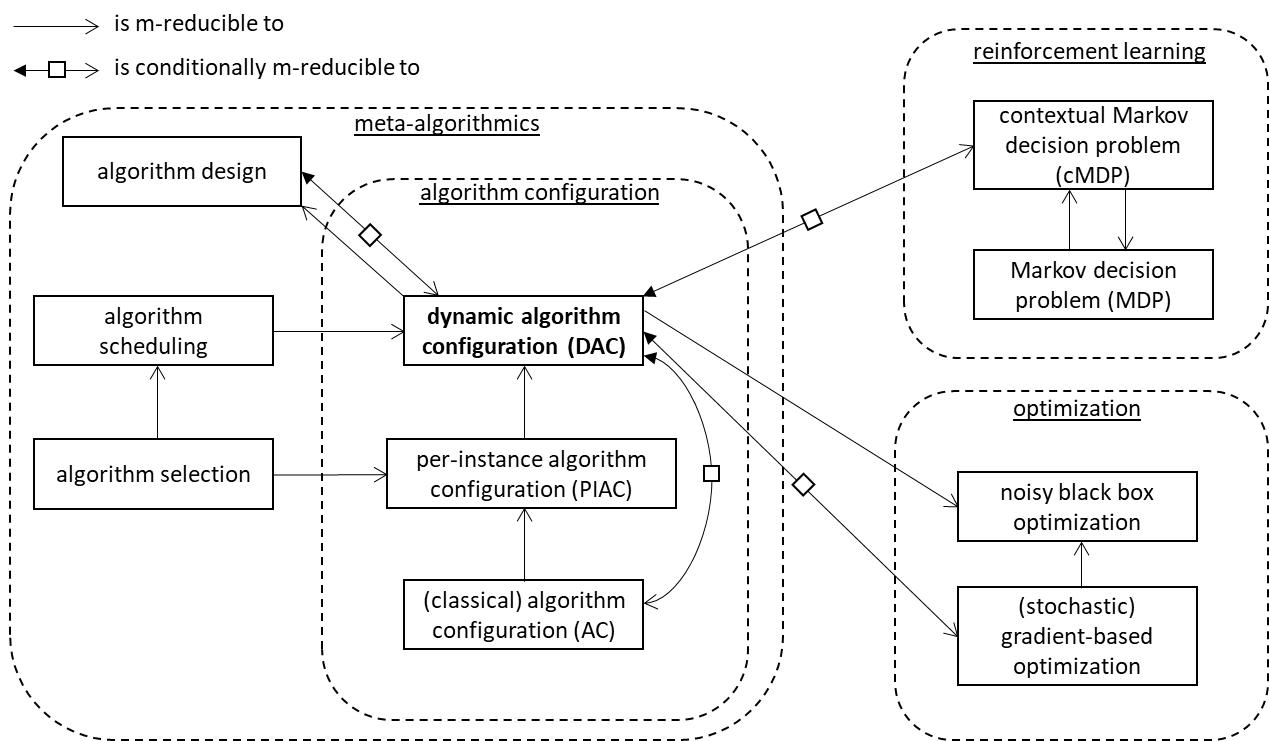}
    \caption{An overview of the reducibility relations between DAC and problems previously studied in the meta-algorithmics, reinforcement learning, and optimization communities; that we prove to exist in Section~\ref{a:proofs}. Note that arrows implied by the transitive/reflexive property of $m$-reducibility are not shown. Conditional reducibility indicates that a problem is only reducible to another, under specific conditions (i.e., not generally).}
    \label{fig:dacreducibility}
\end{figure}

Figure~\ref{fig:dacreducibility} shows an overview of the reducibility relationships between DAC and all the other computational problems we discussed in the main text.

\paragraph{Reducibility to DAC:}
We observe that all problems in meta-algorithmics, discussed in Section~\ref{sec:meta}, can be shown to be generally reducible to DAC. This suggests that research towards solving DAC in general, will indirectly find applications in solving many of these problems. Note that the \emph{conditional} reduction from algorithm design corresponds to the ``DAC powered PbO'', discussed in Section~\ref{sec:aad}, despite being conditional (i.e., not general), presents a highly practical approach to automating algorithm design.

\paragraph{Reducibility from DAC:} We observe that DAC is generally reducible to ``algorithm design'' (as in Definition~\ref{def:adp}). However, this reduction is not practical since no general solvers for this problem are known. DAC is also generally reducible to ``noisy black box optimization''. While this reduction is more practical (see Section~\ref{sec:opt4dac}), a lot of information is lost in the process. Beyond these two problems, DAC is conditionally reducible. While many of these conditional reductions give rise to practical solution approaches (see Section~\ref{sec:methods}), they are nonetheless limited both in terms of generality and the information they can exploit. 

\paragraph{Conclusion:} While a general DAC solver would allow us to solve many well-known computational problems, no such solver exists to date, and is therefore a research direction worth exploring. Note that various existing solvers can solve special cases of DAC, suggesting that another line of research would be to identify further special cases that can be solved more efficiently.


\subsection{Reducibility Proofs}
\label{a:proofs}
In this subsection, we formalize the reducibility relationships between DAC (Definition~\ref{def:dac}) and all problems discussed in the main text. Here, we first formally define each problem and then show reducibility by describing one possible reduction, i.e., defining  $\formulate$ and $\interpret$ functions. Finally, we present a formal argument (\wordasword{proof sketch}) for the correctness of each reduction.\footnote{For brevity, we generally proof (1) in Definition~\ref{def:ms}, but not corner case (2).}

\subsubsection{Algorithm Configuration}
\label{a:ac}
All discussed algorithm configuration variants were already defined in the main text:
\begin{itemize}
\item classical / per-distribution algorithm configuration (AC, Definition~\ref{def:ac})
\item per-instance algorithm configuration (PIAC, Definition~\ref{def:piac})
\item dynamic algorithm configuration (DAC, Definition~\ref{def:dac})
\end{itemize}
In what follows, we formalize their relation.

\paragraph{AC $\leq_m$ PIAC:}
\begin{description}
\item[$\formulate(\langle \mathcal{A}, \Theta, \mathcal{D}, c \rangle)$] $= \langle \mathcal{A}, \Theta, \mathcal{D}, \Psi, c' \rangle$ with
\begin{itemize}
    \item $\Psi = \{\psi_{\vect{\theta}} \, | \, \psi_{\vect{\theta}}(i) = {\vect{\theta}}, \forall {\vect{\theta}} \in \Theta\}$
    \item $c'(\psi_{\vect{\theta}}, i) = c({\vect{\theta}}, i)$
\end{itemize}
\item[$\interpret(\langle \mathcal{A}, \Theta, \mathcal{D}, c \rangle, \psi_{{\vect{\theta}}^*})$] $= {\vect{\theta}}^*$
\item[Proof Sketch:] ${\vect{\theta}}^* \in \argmin_{{\vect{\theta}} \in \Theta} \, \mathbb{E}_{i \sim \mathcal{D}} \, [c({\vect{\theta}}, i)]$ \\ By contradiction, assume ${\vect{\theta}}^* \notin \argmin_{{\vect{\theta}} \in \Theta} \, \mathbb{E}_{i \sim \mathcal{D}} \, [c({\vect{\theta}}, i)]$. This implies that there exists ${\vect{\theta}}' : \mathbb{E}_{i \sim \mathcal{D}}\, [c({\vect{\theta}}', i)] < \mathbb{E}_{i \sim \mathcal{D}}\, [c({\vect{\theta}}^*, i)]$. Since ${\vect{\theta}}' \in \Theta$ and $c({\vect{\theta}}, i) = c'(\psi_{\vect{\theta}}, i)$, there must exist $\psi_{{\vect{\theta}}'} \in \Psi$ having $\mathbb{E}_{i \sim \mathcal{D}}\, [c'(\psi_{{\vect{\theta}}'}, i)] < \mathbb{E}_{i \sim \mathcal{D}}\, [c'({\psi_{{\vect{\theta}}^*}}, i)]$, contradicting $\psi_{{\vect{\theta}}^*} \in \argmin_{\psi \in \Psi} \, \mathbb{E}_{i \sim \mathcal{D}} \, [c'(\psi, i)]$. $\square$
\end{description}

\paragraph{PIAC $\leq_m$ DAC:}
\begin{description}
\item[$\formulate(\langle \mathcal{A}, \Theta, \mathcal{D}, \Psi, c \rangle)$] $= \langle \mathcal{A}', \Theta, \mathcal{D}, \Pi, c' \rangle$ with
\begin{itemize}
    \item $\mathcal{A}'.\step(s, i, {\vect{\theta}}) = \mathcal{A}(i, {\vect{\theta}})$ and $\mathcal{A}'.\init(i) = \upsilon$ and $\mathcal{A}'.\isfinal(s,i) \iff s \neq \upsilon$ (for some $\upsilon$ not being an output of $\mathcal{A}$, i.e., we perform exactly one step)
    \item $\Pi = \{\pi_\psi | \pi_\psi(s, i) = \psi(i), \forall \psi \in \Psi \}$
    \item $c'(\pi_\psi, i) = c(\psi, i)$
\end{itemize}
\item[$\interpret(\langle \mathcal{A}, \Theta, \mathcal{D}, \Psi, c \rangle, \pi_{\psi^*})$] $= \psi^*$
\item[Proof Sketch:] $\psi^* \in \argmin_{\psi \in \Psi} \, \mathbb{E}_{i \sim \mathcal{D}} \, [c(\psi, i)]$. \\ By contradiction, assume $\psi^* \notin \argmin_{\psi \in \Psi} \, \mathbb{E}_{i \sim \mathcal{D}} \, [c(\psi, i)]$. This implies that there exists $\psi' : \mathbb{E}_{i \sim \mathcal{D}}\, [c(\psi', i)] < \mathbb{E}_{i \sim \mathcal{D}}\, [c(\psi^*, i)]$. Since $\psi' \in \Psi$, and $c(\psi, i) = c'(\pi_\psi, i)$, there must exist a $\pi_{\psi'} \in \Pi$ having $\mathbb{E}_{i \sim \mathcal{D}}\, [c'(\pi_{\psi'}, i)] < \mathbb{E}_{i \sim \mathcal{D}}\, [c'({\pi_{\psi^*}}, i)]$, contradicting $\pi_{\psi^*} \in \argmin_{\pi \in \Pi} \, \mathbb{E}_{i \sim \mathcal{D}} \, [c'(\pi, i)]$. $\square$
\end{description}

\paragraph{DAC $\leq^c_m$ AC:}
\begin{description}
\item[Preconditions:] We assume to be given a parametric representation $\Lambda$ of the policy space, i.e., $\Pi = \{\pi_{\vect{\lambda}} \, | \, \vect{\lambda} \in \Lambda \}$. 
\item[$\formulate(\langle \mathcal{A}, \Theta, \mathcal{D}, \Pi, c \rangle)$] $= \langle \mathcal{A}', \Lambda, \mathcal{D}, c' \rangle$ with
\begin{itemize}
    \item $\mathcal{A}'(i,\vect{\lambda}) = \mathcal{A}(i,\pi_{\vect{\lambda}})$.
    \item $c'(\vect{\lambda}, i) = c(\pi_{\vect{\lambda}}, i)$
\end{itemize}
\item[$\interpret(\langle \mathcal{A}, \Theta, \mathcal{D}, \Pi, c \rangle,\vect{\lambda}^*)$] $= \pi_{\vect{\lambda}^*}$ 
\item[Proof Sketch:] $\pi_{\vect{\lambda}^*} \in \argmin_{\pi_{\vect{\lambda}} \in \Pi} \, \mathbb{E}_{i \sim \mathcal{D}} \, [c(\pi_{\vect{\lambda}}, i)]$
\begin{align*} 
\vect{\lambda}^* \in \argmin_{\vect{\lambda} \in \Lambda} \, \mathbb{E}_{i \sim \mathcal{D}} \, [c'(\vect{\lambda}, i)] \implies \\
\vect{\lambda}^* \in \argmin_{\vect{\lambda} \in \Lambda} \, \mathbb{E}_{i \sim \mathcal{D}} \, [c(\pi_{\vect{\lambda}}, i)] \implies \\
\pi_{\vect{\lambda}^*} \in \argmin_{\pi_{\vect{\lambda}} \in \Pi} \, \mathbb{E}_{i \sim \mathcal{D}} \, [c(\pi_{\vect{\lambda}}, i)]
\end{align*}
$\square$
\end{description}

\subsubsection{Algorithm Design}
We formalize algorithm design as in previous work~\shortcite{adriaensen-phd18}:
\begin{definition}[Algorithm Design \label{def:adp}]
Let $A_U$ be the space of all algorithms.\footnote{$A_U$ is a universal set containing “any procedure that solves some problem”. Further formalization of this notion is hindered by the lack of a generally accepted, formal definition of ``an algorithm''.}
Given a preference relation over $\preceq$\footnote{$\preceq$ is assumed to be a preorder, i.e., a binary relation that is reflexive and transitive.} over $A_U$, find a $a^* \in A_U: a^* \nprec a, \forall a \in A_U$.
\end{definition}

\paragraph{algorithm design $\leq^c_m$ DAC:}
\begin{description}
\item[Preconditions:] 
We assume we are given $\init$, $\step$, $\isfinal$, and a set of $\Pi$ sub-routines such that algorithms $a_\pi \in A_{\Pi} \subset A_U$ that can be decompose in $\langle \init, \step, \isfinal, \pi \rangle$ as in Algorithm~\ref{alg:dac} are at least as preferable as any other: ${\forall a_\pi \in A_{\Pi}:  a \preceq a_\pi, \forall a \in A_U \setminus A_{\Pi}}$.

\item[$\formulate(\langle \preceq, \init, \step, \isfinal, \Pi \rangle)$] $= \langle \mathcal{A}, \Theta, \mathcal{D}, \Pi, c \rangle$ with
\begin{itemize}
    \item $\mathcal{A}.\step = \step$ and  $\mathcal{A}.\init = \init$ and $\mathcal{A}.\isfinal = \isfinal$
    \item Any choice of $\mathcal{D}$ and $c$ such that $a_{\pi'} \prec a_\pi \implies \mathbb{E}_{i \sim \mathcal{D}} \, [c(\pi, i)] < \mathbb{E}_{i \sim \mathcal{D}} \, [c(\pi', i)]$. This can always be achieved, as a $c$ that is solely a function of $\pi$ can impose an arbitrary total order on $\Pi$ and therefore also one consistent with $\preceq$.
\end{itemize}
\item[$\interpret(\langle \preceq, \init, \step, \isfinal, \Pi \rangle, \pi^*$)] $= a_{\pi^*}$
\item[Proof Sketch:] $a_{\pi^*} \nprec a, \forall a \in A_U$ \\ By contradiction, assume there exists $a' \in A_U: a' \succ a_{\pi^*}$. Given our pre-condition, we have $a' = a_{\pi'} \in A_{\Pi}$. From our choice of $c$ follows that $\mathbb{E}_{i \sim \mathcal{D}}\, [c(\pi', i)] < \mathbb{E}_{i \sim \mathcal{D}}\, [c(\pi^*, i)]$, contradicting $\pi^* \in \argmin_{\pi \in \Pi} \, \mathbb{E}_{i \sim \mathcal{D}} \, [c(\pi, i)]$. $\square$ 
\end{description}

\paragraph{DAC $\leq_m$ algorithm design:}
\begin{description}
\item[$\formulate(\langle \mathcal{A}, \Theta, \mathcal{D}, \Pi, c \rangle)$] $= \, \preceq$ \hfill \\ satisfying 
\begin{enumerate}
\item $\forall \pi \in \Pi: a \prec \pi, \forall a \in A_U \setminus \Pi$ and
\item $\forall \pi, \pi' \in \Pi: \mathbb{E}_{i \sim \mathcal{D}}\, [c(\pi, i)] < \mathbb{E}_{i \sim \mathcal{D}}\, [c(\pi', i)] \implies \pi' \prec \pi$. 
\end{enumerate}
\item[$\interpret(\langle \mathcal{A}, \Theta, \mathcal{D}, \Pi, c \rangle, a^*$)] $= a^*$
\item[Proof Sketch:] $a^* \in \argmin_{\pi \in \Pi} \, \mathbb{E}_{i \sim \mathcal{D}} \, [c(\pi, i)]$ \\
By contradiction, assume $a^* \notin \argmin_{\pi \in \Pi} \, \mathbb{E}_{i \sim \mathcal{D}} \, [c(\pi, i)]$. First, note that $a^* \in A_U \setminus \Pi$ and (1) would contradict $a^* \in A_U: a^* \nprec a, \forall a \in A_U$. This implies there exists $\pi' \in \Pi: \mathbb{E}_{i \sim \mathcal{D}} \, [c(\pi', i)] < \mathbb{E}_{i \sim \mathcal{D}} \, [c(a^*, i)]$, implying $a^* \prec \pi'$ by (2) and contradicting $a^* \in A_U: a^* \nprec a, \forall a \in A_U$. 
$\square$ 
\end{description}

\subsubsection{Algorithm Selection}
For algorithm selection, we adopt the classical definition by  \shortciteA{rice76a}:\footnote{\shortciteA{rice76a} defines many different more general variants of the problem. However, we will restrict ourselves to a canonical variant, i.e., selecting the best algorithm, per-instance, from finite alternatives, without constraints on the selection mappings.}
\begin{definition}[Algorithm Selection  \label{def:theory_as}]
Given $\langle A, I, c \rangle$:
\begin{itemize}
\setlength{\itemindent}{1em}
\item[--] A finite set $A$ of target algorithms 
\item[--] A target problem space $I$
\item[--] A cost metric $c: A \times I \rightarrow \mathbb{R}$ assessing the cost of solving $i \in I$ using $a \in A$.\footnote{The original definition uses a performance metric $p$ (to be maximized).} 
\end{itemize}
Find a selection mapping $S^*: I \rightarrow A$ satisfying $S^*(i) \in \argmin_{a \in A}\, c(a,i), \forall i \in I$.
\end{definition}
The reducability algorithm selection to DAC follows by transitive property from its reducability to PIAC.

\paragraph{algorithm selection $\leq_m$ PIAC:}
\begin{description}
\item[$\formulate(\langle A, I, c \rangle)$] $= \langle \mathcal{A}', \Theta, \mathcal{D}, \Psi, c' \rangle$ with
\begin{itemize}
    \item $\mathcal{A}'(i, k) = a_k(i)$
    \item $\Theta = \{k \, | \, a_k \in A\}$ (single categorical parameter)
    \item $\mathcal{D} = U(I)$
    \item $\Psi: I \rightarrow \Theta$ (unconstrained)
    \item $c'(\psi, i) = c(\psi(i), i)$
\end{itemize}
\item[$\interpret(\langle A, I, c \rangle, \psi^*)$] $= S^*$ with $S^*(i) = a_{\psi^*(i)}$
\item[Proof Sketch:] $S^*(i) \in \argmin_{a \in A} c(a,i), \forall i \in I$ \\ By contradiction, assume $\exists j \in I: S^*(j) \notin \argmin_{a \in A} c(a,j)$.  This implies there exists $S': S'(j) \in \argmin_{a \in A} c(a,j) \land S'(i') = S(i'), \, \forall i' \in I \setminus \{j\}$. Since $\Psi$ is unconstrained every selection mapping $S$ has its corresponding $\psi \in \Psi: S(i) = a_{\psi(i)}$ and therefore $\exists \, \psi' \in \Psi: c(\psi'(j), j) < c(\psi^*(j), j) \, \land \, c(\psi'(i'), i') = c(\psi^*(i), i'), \, \forall i' \in I \setminus \{j\}$. From $c'(\psi, i) = c(\psi(i), i)$ and $\mathcal{D}(j) > 0$ follows that $\mathbb{E}_{i \sim \mathcal{D}}\, [c'(\psi', i)] < \mathbb{E}_{i \sim \mathcal{D}}\, [c'(\psi^*, i)]$, contradicting $\psi^* \in \argmin_{\psi \in \Psi} \, \mathbb{E}_{i \sim \mathcal{D}} \, [c'(\psi, i)]$. $\square$
\end{description}

\subsubsection{Algorithm Scheduling}
We define a variant of algorithm scheduling that considers allocating a fixed time budget to a finite set of target algorithms in an instance-aware and dynamic fashion:
\begin{definition}[Algorithm Scheduling  \label{def:as}]
Given $\langle A, B, I, \Delta, c \rangle$:
\begin{itemize}
\setlength{\itemindent}{1em}
\item[--] A finite set $A$ of step-wise executable target algorithms such that the execution of the $k^{\textrm{th}}$ algorithm $a_k \in A$ can be decomposed as a consecutive application of a sub-routine $a_k.\timestep$ such that the state of algorithm $a_k$ when solving a problem instance $i$, after $t$ time steps, is given by $s_{k,i,t} = a_k.\timestep(s_{k,i,t-1})$ with $s_{k,i,0}=i$.
\item[--] A finite budget $B$ of time steps to be allocated to algorithms in $A$.
\item[--] A target problem space $I$
\item[--] A space of \emph{dynamic scheduling policies} $\delta \in \Delta$ with $\delta: \mathcal{S}^{|A|} \times I \times \mathbb{N} \rightarrow A$ 
choosing which algorithm $a_k \in A$ to resume executing in the next time step, as a function of the total time $T \in \mathbb{N}$ elapsed thus far, the instance $i \in I$ being solved and the vector of states $s_{m,i,t_m}$ of each algorithm $a_m \in A$.
\item[--] A cost metric $c: A \times I \times \mathbb{N}  \rightarrow \mathbb{R}$ assessing the cost of solving $i \in I$ using $a \in A$ for $t \in \mathbb{N}$ time steps.
\end{itemize}
Find a dynamic scheduling policy $\delta^* \in \argmin_{\delta \in \Delta}\, \sum_{i \in I} (\min_{a_k \in A} c(a_k,i,t_k^{B,\delta,i}))$ where $t_k^{T,\delta,i}$ is the total time allocated to $a_k$ by $\delta$ after $T$ scheduling steps on instance $i$, and is given by \newline $t_k^{T,\delta, i} = \begin{cases}
0 & T = 0 \\
t_k^{T-1,\delta, i} &  T > 0 \quad \land \quad a_k \neq \delta(\vect{s}^{T-1,\delta,i},i,T-1) \\
t_k^{T-1,\delta, i} + 1 & T > 0 \quad \land \quad  a_k = \delta(\vect{s}^{T-1,\delta,i},i,T-1)
\end{cases}$ \newline with \newline $s_k^{T,\delta, i} = \begin{cases}
i & T = 0 \\
s_k^{T-1,\delta, i} &  T > 0 \quad \land \quad a_k \neq \delta(\vect{s}^{T-1,\delta,i},i,T-1) \\
a_k.\timestep(s_k^{T-1,\delta, i}) & T > 0 \quad \land \quad  a_k = \delta(\vect{s}^{T-1,\delta,i},i,T-1)
\end{cases}$
\end{definition}

\paragraph{algorithm scheduling $\leq_m$ DAC:}
\begin{description}
\item[$\formulate(\langle A, B, I, \Delta, c \rangle)$] $= \langle \mathcal{A}, \Theta, \mathcal{D}, \Pi, c' \rangle$ with
\begin{itemize}
    \item $\mathcal{A}.\step((\vect{s},i,T),i,\theta) = (\vect{s}',i,T+1)$ with $s'_k = \begin{cases} 
    s_k & \theta \neq k \\
    a_k.\timestep(s_k) & \theta = k
    \end{cases}$ \newline $\mathcal{A}.\init(i) = (\vect{s},i,0)$ with $s_k = i$ \newline $\mathcal{A}.\isfinal((\vect{s},i,T),i) \Leftrightarrow T = B$.
    \item $\Theta = \{k \, | \, a_k \in A\}$
    \item $\mathcal{D} = U(I)$
    \item $\Pi = \{\pi_\delta \, | \, \delta \in \Delta\}$ with $\pi_\delta((\vect{s},i,T),i) = k \Leftrightarrow \delta(\vect{s},i,T) = a_k$
    \item $c'(\pi_\delta, i) = \min_{a_k \in A} c(a_k,i,t_k^{B,\delta,i})$
\end{itemize}
\item[$\interpret(\langle A, B, I, \Delta, c \rangle, \pi_{\delta^*})$] $= \delta^*$
\item[Proof Sketch:] $\delta^* \in \argmin_{\delta \in \Delta}\, \sum_{i \in I} (\min_{a_k \in A} c(a_k,i,t_k^{B,\delta,i}))$
\begin{align*} 
\pi_{\delta^*} \in \argmin_{\pi_\delta \in \Pi} \, \mathbb{E}_{i \sim \mathcal{D}} \, [c'(\pi_\delta, i)] \implies \\
\delta^* \in \argmin_{\delta \in \Delta} \, \mathbb{E}_{i \sim \mathcal{D}} \, [\min_{a_k \in A} c(a_k,i,t_k^{B,\delta,i})] \implies \\
\delta^* \in \argmin_{\delta \in \Delta} \, \frac{1}{|I|} \sum_{i \in I} \, [\min_{a_k \in A} c(a_k,i,t_k^{B,\delta,i})] \implies \\
\delta^* \in \argmin_{\delta \in \Delta}\, \sum_{i \in I} (\min_{a_k \in A} c(a_k,i,t_k^{B,\delta,i}))
\end{align*}
$\square$
\end{description}
\vspace{5.5cm}
\subsubsection{Reinforcement Learning}
\label{a:rl}
In Section~\ref{sec:rl4dac}, we discussed the relation between DAC and the 
\begin{definition}[Markov Decision Problem (MDP) \label{def:mdp}]
Given $\langle S, A, T, R \rangle$:
\begin{itemize}
\setlength{\itemindent}{1em}
\item[--] A state space $S$.
\item[--] An action space $A$
\item[--] A transition function $T: S \times A \rightarrow S$
\item[--] A reward function $R: S \times A \rightarrow \mathbb{R}$
\end{itemize}
Find a policy $\pi: S \rightarrow A$ satisfying $\pi^*(s) \in \argmax_{a \in A} R(s, a) + V^*(T(s,a))$ with $V^*$ being the optimal value-state function, i.e., $V^*(s) = \max_{a \in A} R(s, a) + V^*(T(s,a))$.
\end{definition}
Note that we will restrict ourselves to episodic MDPs, where we have absorbing states ${S_{\mathcal{H}} = \{ s \, | \, T(s,a) = s, \forall a \in A\}}$ having $R(s,a) = 0, \forall s \in S_\mathcal{H}$ that are reached within an arbitrarily large, but finite horizon $\mathcal{H}$ and therefore $V^*(s) = \max_{\pi} \sum_{t=0}^{\mathcal{H}-1} R(s_{\pi,t},\pi(s_{\pi,t}))$ where $s_{\pi,t} = T(s_{\pi,t-1}, \pi(s_{\pi,t-1}))$ is the $t^\textrm{th}$ state encountered when following $\pi$ starting in $s_{\pi,0} = s$. 

Since standard RL methods are not instance-aware, \shortciteA{biedenkapp-ecai20a} proposed to model DAC as a
\begin{definition}[Contextual Markov Decision Problem \shortcite<cMDP, >{hallak-corr15} \label{def:cmdp}]
Given $\langle \mathcal{C}, S, A, \mathcal{M} \rangle$: 
\begin{itemize}
\setlength{\itemindent}{1em}
\item[--] A context space $\mathcal{C}$
\item[--] A shared state space $S$
\item[--] A shared action space $A$
\item[--] A function $\mathcal{M}$ mapping any $c \in \mathcal{C}$ to an MDP $\mathcal{M}(c) = \langle S, A, T_c, R_c \rangle$ with
\begin{itemize}
\item[--] a context-dependent transition function $T_c: S \times A \rightarrow S$
\item[--] a context-dependent reward function $R_c: S \times A \rightarrow \mathbb{R}$
\end{itemize}
\end{itemize}
Find a policy $\pi: S \times \mathcal{C} \rightarrow A$ satisfying $\pi^*(s, c) \in \argmax_{a \in A}  \, R_c(s, a) + V_c^*(T_c(s,a))$ with $V_c^*(s) = \max_{a \in A}  R_c(s, a) + V_c^*(T_c(s,a))$.
\end{definition}

Please remark that we assume the context to be observable and our objective to be finding an optimal \emph{context-dependent} policy. We will also assume MDP $\mathcal{M}(i)$ to be episodic. As a consequence, this formulation is $m$-equivalent to that of an ordinary episodic MDP. The cMDP formulation is nonetheless interesting in that it can capture various aspects of DAC abstracted in the MDP reduction: DAC $\leq_m$ MDP $\leq_m$ cMDP.

\paragraph{DAC $\leq^c_m$ cMDP:}
\begin{description}
\item[Preconditions:] \hfill
\begin{enumerate}
    \item The cost function $c$ is step-wise decomposable, i.e., we are given functions $\langle \initcost, \stepcost \rangle$, such that 
\begin{align*}
 c(\pi, i) = \initcost(i) + \sum_{t=0}^{T-1} \stepcost(s_t, i, \pi(s_t, i))
\end{align*}
where
\begin{align*}
s_0 = \init(i) \quad \land \quad s_t = \step(s_{t-1}, i, \pi(s_{t-1}, i)) \quad \land \quad \isfinal(s_t, i) \Leftrightarrow t = T
\end{align*}
\item The policy space is unconstrained, i.e., $\Pi = \{\pi \, | \, \pi(s, i) \in \Theta,\, \, \forall \, s \in \mathcal{S} \, \land \, i \in I\}$
\end{enumerate}
\item[$\formulate(\langle \mathcal{A}, \Theta, \mathcal{D}, \Pi, c \rangle)$] $= \langle \mathcal{C}, S, A, \mathcal{M} \rangle$ with
\begin{itemize}
    \item $\mathcal{C} = I$ the domain of $\mathcal{D}$.
    \item $S = \mathcal{S}$ the set of algorithm states.
    \item $A = \Theta$
    \item $\mathcal{M}(i) = \langle S, A, T_i, R_i \rangle$ with
\begin{itemize}
\item[--] $T_i(s, \vect{\theta}) = \begin{cases}
s & \isfinal(s, i)\\
\step(s, i, \vect{\theta})& \lnot \isfinal(s, i)
\end{cases}$
\item[--] $R_i(s, \vect{\theta}) = \begin{cases}
0 & \isfinal(s, i)\\
\stepcost(s, i, \vect{\theta})& \lnot \isfinal(s, i)
\end{cases}$
\end{itemize}
\end{itemize}
\item[$\interpret(\langle \mathcal{A}, \Theta, \mathcal{D}, \Pi, c \rangle, \pi^*)$] $= \pi^*$ 
\item[Proof Sketch:] $\pi^* \in \argmax_{\pi}  \, \mathbb{E}_{i \sim \mathcal{D}} \, c(\pi, i)$ \\ 
We first note that since $S = \mathcal{S} \, \land \, A = \Theta \, \land \, \mathcal{C} = I$ and given precondition (2), it follows that both problems have the exact same policy space $\Pi$. It remains to show that optimality in the resulting cMDP implies optimality in the original DAC:
\begin{align*} 
\pi^*(s, i) \in \argmax_{\vect{\theta} \in \Theta}  \, R_i(s, \vect{\theta}) + V_i^*(T_i(s,\vect{\theta})) \quad (\forall s \in S, \forall i \in I) \implies \\
\pi^*(\init(i), i) \in \argmax_{\vect{\theta} \in \Theta}  \, R_i(\init(i), \vect{\theta}) + V_i^*(T_i(\init(i),\vect{\theta})) \quad (\forall i \in I)  \implies \\
\pi^* \in \argmax_{\pi}  \, \mathbb{E}_{i \sim \mathcal{D}} \, R_i(\init(i), \pi(\init(i),i)) + V_i^*(T_i(\init(i),\pi(\init(i),i)))
\end{align*}
Since each $\mathcal{M}(i)$ is episodic, this objective can be rewritten as:
\begin{gather*} 
\pi^* \in \argmax_{\pi}  \, \mathbb{E}_{i \sim \mathcal{D}} \sum_{t=0}^{\mathcal{H}-1} R_i(s_{\pi,t,i},\pi(s_{\pi,t,i})) \text{ where } s_{\pi, t, i} = \begin{cases}
\init(i) & t = 0 \\
T_i(s_{\pi,t-1,i}, \pi(s_{\pi,t-1,i})) & t > 0 \end{cases} \\
\pi^* \in \argmax_{\pi}  \, \mathbb{E}_{i \sim \mathcal{D}} \sum_{t=0}^{T-1} \stepcost(s_{\pi,t,i},i,\pi(s_{\pi,t,i})) \text{ where } s_{\pi, t, i} = \begin{cases}
\init(i) & t = 0 \\
\step(s_{\pi,t-1,i}, i, \pi(s_{\pi,t-1,i})) & t > 0
\end{cases} \\
\pi^* \in \argmax_{\pi}  \, \mathbb{E}_{i \sim \mathcal{D}} \, c(\pi, i)
\end{gather*}
$\square$
\end{description}
It is worth noting that the optimality of $\pi^*$ does not depend on $\mathcal{D}$, $\init$, or $\initcost$.

\noindent Conditional reducability to an ordinary MDP follows from the transitive property and
\paragraph{cMDP $\leq_m$ MDP:}
\begin{description}
\item[$\formulate(\langle \mathcal{C}, S, A, \mathcal{M} \rangle)$] $= \langle S', A, T, R \rangle$ with
\begin{itemize}
    \item $S' = S \times \mathcal{C}$
    \item $T((s,c),a) = T_c(s,a)$
    \item $R((s,c),a) = R_c(s,a)$
\end{itemize}
where $\mathcal{M}(c) = \langle S, A, T_c, R_c \rangle$.
\item[$\interpret(\langle \mathcal{C}, S, A, \mathcal{M} \rangle,\pi^*)$] $= \pi'^*$ with $\pi'^*(s,c) = \pi^*((s,c))$
\item[Proof Sketch:] $\pi'^*(s,c) \in \argmax_{a \in A} R_c(s, a) + V_c^*(s)$ \\
\begin{align*} 
\pi^*((s,c)) \in \argmax_{a \in A} R((s,c), a) + V^*(T((s,c),a)) \implies \\
\pi'^*(s,c) \in \argmax_{a \in A} R_c(s, a) + V^*(T_c(s,a)) \\
\end{align*}
It remains to show that $V^*((s,c)) = V_c^*(s)$. We can proof this by induction on the maximum number of steps $\mathcal{H}_{s,c}$ before reaching an absorbing state following any policy starting from state $s$ in context $c$. The base case $\mathcal{H}_{s,c}=0$ follows from ${T((s,c), a) = (s, c)} \iff T_c(s,a) = s$. Now, for the recursive case, assuming this holds for $\mathcal{H}_{s,c} \leq n-1$, it also holds for $\mathcal{H}_{s,c} = n$
\begin{align*} 
V^*((s,c)) &= \max_{a \in A} R((s,c), a) + V^*(T((s,c),a)) \\
&= \max_{a \in A} R_c(s, a) + V^*(T_c(s,a)) \\
&= \max_{a \in A} R_c(s, a) + V_c^*(T_c(s,a)) = V_c^*(s)
\end{align*}
since $\mathcal{H}_{T_c(s,a),c} \leq n-1$.
$\square$
\end{description}
and both are $m$-equivalent since
\paragraph{MDP $\leq_m$ cMDP:}
\begin{description}
\item[$\formulate(\langle S, A, T, R \rangle)$] $= \langle \mathcal{C}, S, A, \mathcal{M} \rangle$ with
\begin{itemize}
    \item $\mathcal{C} = \{0\}$
    \item $T_0(s, a) = T(s,a)$
    \item $R_0(s, a) = R(s,a)$
\end{itemize}
where $\mathcal{M}(0) = \langle S, A, T_0, R_0 \rangle$.
\item[$\interpret(\langle S, A, T, R \rangle,\pi^*)$] $=\pi'^*$ with $\pi'^*(s) = \pi^*(s,0)$ 
\item[Proof Sketch:] $\pi'^*(s) \in \argmax_{a \in A}  \, R(s, a) + V^*(s)$ \\
\begin{align*} 
\pi^*(s, 0) \in \argmax_{a \in A}  \, R_0(s, a) + V_0^*(T_0(s,a)) \implies \\
\pi'^*(s) \in \argmax_{a \in A}  \, R(s, a) + V_0^*(T(s,a)) \\
\end{align*}

\pagebreak

It remains to show that $V^*(s) = V_0^*(s)$. We can proof this by induction on the maximum number of steps $\mathcal{H}_s$ before reaching an absorbing state following any policy starting from state $s$. The base case $\mathcal{H}_s = 0$ follows from $T(s, a) = s \Leftrightarrow T_0(s,a) = s$. Now the recursive case, assuming this holds for $\mathcal{H}_s \leq n-1$, it also holds for $\mathcal{H}_s = n$
\begin{align*} 
V^*(s) &= \max_{a \in A} R(s, a) + V^*(T(s,a)) \\
&= \max_{a \in A} R_0(s, a) + V^*(T_0(s,a)) \\
&= \max_{a \in A} R_0(s, a) + V_0^*(T_0(s,a)) = V_0^*(s)
\end{align*}
since $\mathcal{H}_{T_0(s,a)} \leq n-1$.
$\square$
\end{description}

\subsubsection{Optimization}
\label{a:opt}
In Section~\ref{sec:opt4dac}, we discussed the relation between DAC and
\begin{definition}[Noisy Black Box Optimization \label{def:bbopt}]
Given $\langle X, e \rangle$:
\begin{itemize}
\setlength{\itemindent}{1em}
\item[--] A search space $X$ 
\item[--] A noisy evaluation sub-routine $e$
\end{itemize}
Find a $x^* \in \argmin_{x \in X}\, \mathbb{E}[e(x)]$.
\end{definition}

\paragraph{DAC $\leq_m$ noisy black box optimization:}
\begin{description}
\item[$\formulate(\langle \mathcal{A}, \Theta, \mathcal{D}, \Pi, c \rangle)$] $= \langle \Pi, e \rangle$ where $e(\pi) = c(\pi, i)$ with $i \sim \mathcal{D}$.
\item[$\interpret(\langle \mathcal{A}, \Theta, \mathcal{D}, \Pi, c \rangle),\pi^*)$] $= \pi^*$
\item[Proof Sketch:] $\pi^* \in \argmin_{\pi \in \Pi} \, \mathbb{E}_{i \sim \mathcal{D}} \, [c(\pi, i)]$ \\
We have $\pi^* \in \argmin_{\pi \in \Pi}\, \mathbb{E}[e(\pi)]$, since $\pi^*$ is a solution for noisy black box optimization problem. Since $e(\pi) = c(\pi, i)$ and $i \sim \mathcal{D}$,  this is equivalent to ${\pi^* \in \argmin_{\pi \in \Pi} \, \mathbb{E}_{i \sim \mathcal{D}} \, [c(\pi, i)]}$. 
$\square$
\end{description}

\noindent Also in Section~\ref{sec:opt4dac}, we discussed the possibility of solving DAC using

\begin{definition}[Stochastic Gradient-Based Optimization \label{def:sgdopt}]
Given $\langle X, e, e' \rangle$:
\begin{itemize}
\item[--] $X$ and $e$ as in Definition~\ref{def:bbopt}.
\item[--] A stochastic differentiation sub-routine $e'$ satisfying $\mathbb{E}[e'(x)] = \mathbb{E}[\frac{\partial e(x)}{\partial x}]$.
\end{itemize}
Find a $x^* \in \argmin_{x \in X}\, \mathbb{E}[e(x)]$.
\end{definition}
\paragraph{DAC $\leq^c_m$ stochastic gradient-based optimization:}
\begin{description}
\item[Preconditions:] \hfill
\begin{enumerate}
\item We assume to be given a parametric representation $\Lambda$ of the policy space, i.e., $\Pi = \{\pi_{\vect{\lambda}} \, | \, \vect{\lambda} \in \Lambda \}$.
\item We assume $c(\pi_{\vect{\lambda}},i)$ to be piece-wise differentiable w.r.t. $\vect{\lambda}$ and a sub-routine for calculating $\frac{\partial c(\pi_{\vect{\lambda}},i)}{\partial \vect{\lambda}}$ to be given.
\end{enumerate}
\item[$\formulate(\langle \mathcal{A}, \Theta, \mathcal{D}, \Pi, c \rangle)$] $= \langle \Lambda, e, e' \rangle$ with
\begin{itemize}
    \item $e(\vect{\lambda}) = c(\pi_{\vect{\lambda}}, i)$ with $i \sim \mathcal{D}$.
    \item $e'(\vect{\lambda}) = \frac{\partial c(\pi_{\vect{\lambda}},i)}{\partial \vect{\lambda}}$ with $i \sim \mathcal{D}$.
\end{itemize}
\item[$\interpret(\langle \mathcal{A}, \Theta, \mathcal{D}, \Pi, c \rangle,\vect{\lambda}^*)$]$= \pi_{\vect{\lambda}^*}$
\item[Proof Sketch:] $\pi_{\vect{\lambda}^*} \in \argmin_{\pi_{\vect{\lambda}} \in \Pi} \, \mathbb{E}_{i \sim \mathcal{D}} \, [c(\pi_{\vect{\lambda}}, i)]$ \\ We have $\vect{\lambda}^* \in \argmin_{\vect{\lambda} \in \Lambda}\, \mathbb{E}[e(\vect{\lambda})]$, since $\vect{\lambda}^*$ is a solution for black box optimization problem. Since $e(\vect{\lambda}) = c(\pi_{\vect{\lambda}}, i)$ and $i \sim \mathcal{D}$,  this is equivalent to $\vect{\lambda}^* \in \argmin_{\vect{\lambda} \in \Lambda} \, \mathbb{E}_{i \sim \mathcal{D}} \, [c(\pi_{\vect{\lambda}}, i)]$. Substituting every $\vect{\lambda}$ by its corresponding policy $\pi_{\vect{\lambda}}$, we get $\pi_{\vect{\lambda}^*} \in \argmin_{\pi_{\vect{\lambda}} \in \Pi} \, \mathbb{E}_{i \sim \mathcal{D}} \, [c(\pi_{\vect{\lambda}}, i)]$
$\square$
\end{description}
At first sight, precondition (2) may seem very strong. In what follows, we show a sufficient condition that is arguably not so strong.
\begin{description}
\item[Sufficient Conditions:] Next to the precondition (1), we assume 
\begin{enumerate}
\setcounter{enumi}{2}
    \item to be given a parametric representation of the state space.
    \item the cost function $c$ to be step-wise decomposable in $\langle \initcost, \stepcost \rangle$ such that \begin{align*}
 c(\pi_{\vect{\lambda}}, i) = \initcost(i) + \sum_{t=0}^{T-1} \stepcost(s_t, i, \pi_{\vect{\lambda}}(s_t, i))
\end{align*}
where
\begin{align*}
s_0 = \init(i) \quad \land \quad s_t = \step(s_{t-1}, i, \pi_{\vect{\lambda}}(s_{t-1}, i)) \quad \land \quad \isfinal(s_t, i) \Leftrightarrow t = T
\end{align*} 
\item to be given sub-routines for calculating the following partial derivatives: $\frac{\partial \pi_{\vect{\lambda}}(s, i)}{\partial \vect{\lambda}}$, $\frac{\partial \stepcost(s, i, \vect{\theta})}{\partial \vect{\theta}}$, $\frac{\partial \stepcost(s, i, \vect{\theta})}{\partial s}$, $\frac{\partial \step(s, i, \vect{\theta})}{\partial \vect{\theta}}$, and $\frac{\partial \step(s, i, \vect{\theta})}{\partial s}$.
\end{enumerate}
\item[Proof Sketch:] Under these conditions precondition 2 also holds. \\ The piece-wise derivative of $c$ w.r.t. $\vect{\lambda}$ can be calculated as follows 
\begin{align*}
\frac{\partial c(\pi_{\vect{\lambda}},i)}{\partial \vect{\lambda}} =& 
\frac{\partial \left( \initcost(i) + \sum_{t=0}^{T-1} \stepcost(s_t, i, \pi_{\vect{\lambda}}(s_t, i))\right)}{\partial \vect{\lambda}} \\
=& 
\frac{\partial \initcost(i)}{\partial \vect{\lambda}} + \sum_{t=0}^{T-1} \frac{\partial \stepcost(s_t, i, \pi_{\vect{\lambda}}(s_t, i))}{\partial \vect{\lambda}} \\ 
=& \sum_{t=0}^{T-1} \frac{\partial \stepcost(s_t, i, \pi_{\vect{\lambda}}(s_t, i))}{\partial \vect{\lambda}}
\end{align*}
where 
\begin{align*}
\frac{\partial \stepcost(s_t, i, \pi_{\vect{\lambda}}(s_t, i))}{\partial \vect{\lambda}} =& \frac{\partial \stepcost(s_t, i, \pi_{\vect{\lambda}}(s_t, i))}{\partial \pi_{\vect{\lambda}}(s_t, i)} \cdot \frac{\partial \pi_{\vect{\lambda}}(s_t, i)}{\partial \vect{\lambda}} \\ 
&+ \frac{\partial s_t}{\partial  \vect{\lambda}} \cdot \left( \frac{\partial \stepcost(s_t, i, \pi_{\vect{\lambda}}(s_t, i))}{\partial s_t} + \frac{\partial \stepcost(s_t, i, \pi_{\vect{\lambda}}(s_t, i))}{\partial \pi_{\vect{\lambda}}(s_t, i)} \cdot \frac{\partial \pi_{\vect{\lambda}}(s_t, i)}{\partial s_t} \right)
\end{align*}
with
\begin{align*}
\frac{\partial s_t}{\partial  \vect{\lambda}} &= \frac{\partial \step(s_{t-1}, i, \pi_{\vect{\lambda}}(s_{t-1}, i))}{\pi_{\vect{\lambda}}(s_{t-1}, i)} \cdot \frac{\partial \pi_{\vect{\lambda}}(s_{t-1}, i)}{\partial \vect{\lambda}} \\ 
&+  \frac{\partial s_{t-1}}{\partial  \vect{\lambda}} \cdot \left( \frac{\partial \step(s_{t-1}, i, \pi_{\vect{\lambda}}(s_{t-1}, i))}{\partial s_{t-1}} + \frac{\partial \step(s_{t-1}, i, \pi_{\vect{\lambda}}(s_{t-1}, i))}{\partial \pi_{\vect{\lambda}}(s_{t-1}, i)} \cdot \frac{\partial \pi_{\vect{\lambda}}(s_{t-1}, i)}{\partial s_{t-1}} \right) 
\end{align*}
if $t > 0$ and $\frac{\partial s_0}{\partial  \vect{\lambda}} = 0$ \\

Note that we can reuse the quantities calculated in the previous step, resulting in a procedure known as \emph{forward-mode} differentiation. While we will not derive the formulas here, $\frac{\partial c(\pi_{\vect{\lambda}},i)}{\partial \vect{\lambda}}$ can also be calculated using \emph{reverse-mode} differentiation, a procedure also known as \emph{backpropagation} in the context of neural networks.
$\square$
\end{description}

\end{document}